\documentclass{article}
\usepackage{iclr2026_conference}
\iclrfinalcopy
\usepackage{times}
\usepackage{latexsym}
\usepackage[T1]{fontenc}
\usepackage[utf8]{inputenc}
\usepackage{microtype}
\usepackage{inconsolata}
\usepackage{listings}
\usepackage{graphicx}
\usepackage{placeins}
\usepackage{xcolor}
\usepackage{tcolorbox}
\usepackage{colortbl}
\usepackage{amsmath,amssymb}
\usepackage{booktabs,tabularx}
\usepackage{multirow}
\usepackage{algorithm}
\usepackage{algpseudocode}
\usepackage{wrapfig}
\usepackage{capt-of}
\usepackage{needspace}
\makeatletter

\newcommand{\dreamwrapclear}{%
  \par
  \ifnum\c@WF@wrappedlines>\@ne
    \@tempdima\baselineskip
    \multiply\@tempdima by \numexpr\c@WF@wrappedlines-1\relax
    \vspace{\@tempdima}%
  \fi
  \WFclear
}
\newcommand{\dreamwrapspace}{%
  \par\Needspace{\dimexpr\WF@size-\baselineskip\relax}%
  \suppressfloats[t]%
}
\makeatother
\AddToHook{env/wrapfigure/before}{\dreamwrapclear}
\AddToHook{env/wraptable/before}{\dreamwrapclear}
\AddToHook{env/wrapfigure/after}{\dreamwrapspace}
\AddToHook{env/wraptable/after}{\dreamwrapspace}
\usepackage{hyperref}
\usepackage{url}
\newif\ifshowbiburls
\showbiburlsfalse
\hypersetup{hidelinks,
  pdftitle={Dream-COP: Guiding Algorithm Discovery for Combinatorial Optimization Problems in an Experience World},
  pdfauthor={Kuan Xu, Jianming Wang, Ni Zhang, Yaoxin Wu, Yong Wang, Zhiguang Cao}}
\lstdefinestyle{dreamprogram}{
  language=Python,
  basicstyle=\ttfamily\fontsize{8.5}{10}\selectfont,
  keywordstyle=\color{blue!65!black},
  commentstyle=\color{green!35!black},
  stringstyle=\color{orange!55!black},
  numbers=left,
  numberstyle=\tiny\color{black!45},
  numbersep=8pt,
  stepnumber=1,
  frame=single,
  rulecolor=\color{black!25},
  backgroundcolor=\color{black!2},
  framesep=5pt,
  xleftmargin=2em,
  xrightmargin=0.5em,
  columns=fullflexible,
  keepspaces=true,
  showstringspaces=false,
  breaklines=true,
  breakatwhitespace=true,
  breakautoindent=true,
  breakindent=1em,
  tabsize=4,
  captionpos=t,
  aboveskip=0.7em,
  belowskip=1em
}
\newcommand{\method}{Dream-COP}
\newcommand{\fixed}{\textsc{Fixed}}
\newcommand{\dream}{\textsc{Dream}}
\newcommand{\hist}{\mathcal{H}}
\newcommand{\dev}{\mathcal{D}_{\mathrm{dev}}}
\title{\normalfont\fontsize{15}{18}\selectfont Dream-COP: Guiding Algorithm Discovery for Combinatorial \\
Optimization Problems in an Experience World}
\author{\textbf{Kuan Xu}$^{1}$\thanks{Email: \texttt{kuanxu@smu.edu.sg}.} \quad \textbf{Jianming Wang}$^{2}$ \quad \textbf{Ni Zhang}$^{1}$ \quad
\textbf{Yaoxin Wu}$^{3}$ \quad \textbf{Yong Wang}$^{4}$ \quad \textbf{Zhiguang Cao}$^{1}$ \\[0.5em]
\normalfont $^{1}$Singapore Management University \\
\normalfont $^{2}$University of Science and Technology of China \\
\normalfont $^{3}$Eindhoven University of Technology \\
\normalfont $^{4}$Central South University
}
\begin{document}
\maketitle

\fancyhead{} \renewcommand{\headrulewidth}{0pt}
\begin{abstract}
Large language models (LLMs) enable algorithm discovery for combinatorial optimization problems (COPs) through iterative program generation and evaluation. A central challenge is how to organize search to discover better algorithms within a limited budget. Our key insight is that past search trajectories and their recorded outcomes form an experience world, which can be used to facilitate the COP solving through replay. We introduce Dream-COP, in which a \emph{policy agent} learns executable policies through \emph{dreaming} in this experience world to guide a \emph{discovery agent} in generating and refining COP programs. Specifically, policy replay provides feedback on the quality of the programs revealed and the search steps used, to guide policy selection more effectively. Across five COPs, \textsc{Dream} discovers better algorithms than fixed exploration in both heuristic component design and complete solver generation. Particularly, it matches the optimal values on KP and reduces the gap on OBPP from 2.112\% to 1.199\% while using 42.5\% fewer search requests. Moreover, the learned exploration policies transfer effectively across different COPs with the assitance of the experience world.
\end{abstract}

\section{Introduction}

\begin{wrapfigure}{r}{0.48\textwidth}
  \centering
  \includegraphics[width=\linewidth]{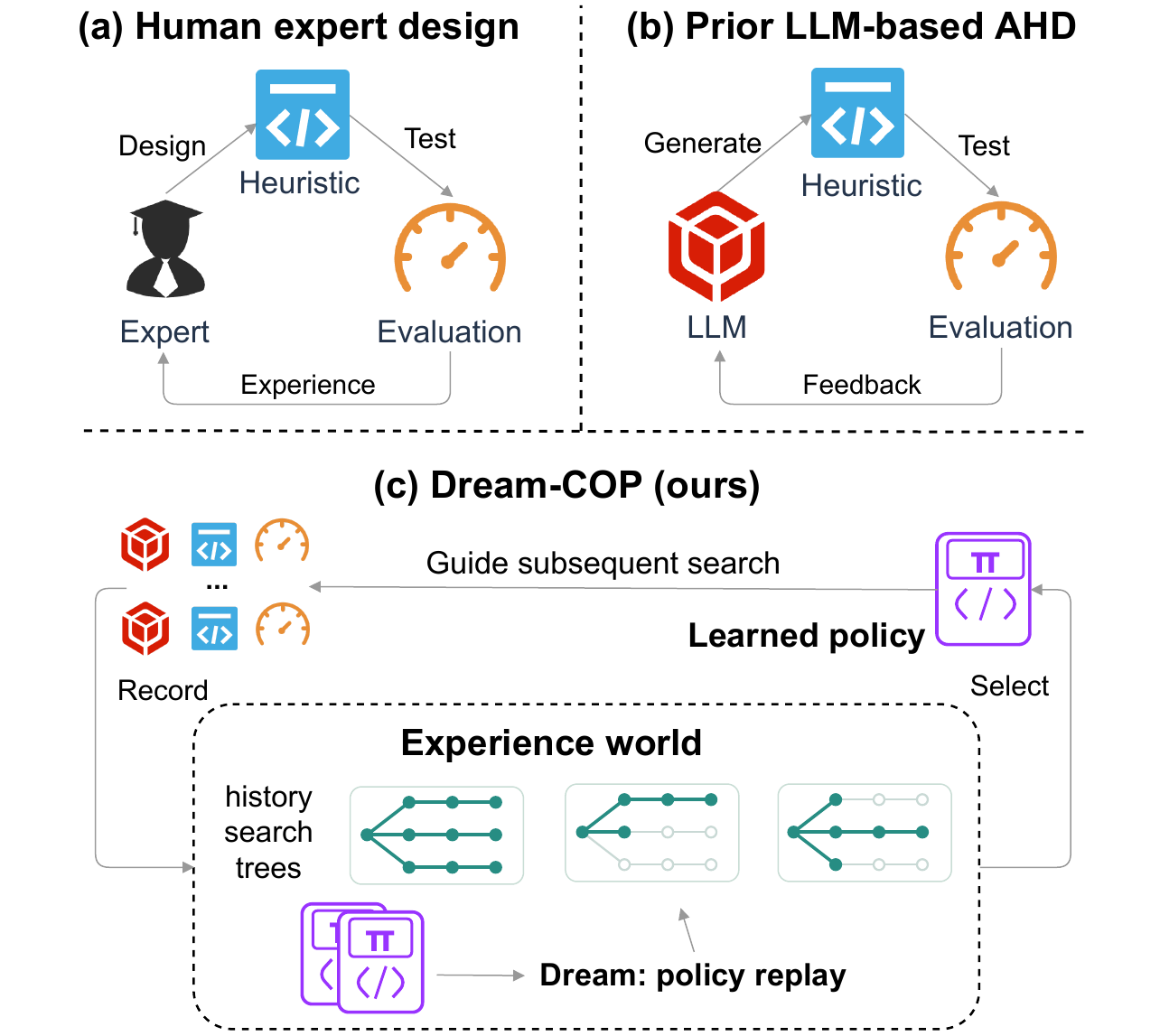}
\caption{Three approaches to heuristic design: (a) human expertise, (b) LLM-based heuristic search, and (c) \method{}, which learns exploration policies through historical replay in the experience world.}
  \label{fig:intro-paradigms}
\end{wrapfigure}

Combinatorial optimization problems (COPs) are central to efficient resource use in transportation, production scheduling, and resource allocation, yet finding optimal solutions becomes increasingly computationally demanding as problem size and complexity grow \citep{korte2018combinatorial}. Heuristics and metaheuristics play a central role in obtaining high-quality solutions under practical computational constraints \citep{gendreau2019handbook}. Developing effective heuristics often requires substantial expertise and repeated experimentation \citep{bengio2021combinatorial}, motivating methods that can automate algorithm design for different problems (Figure~\ref{fig:intro-paradigms}).

Automatic heuristic design (AHD) automates the selection, composition, and configuration of heuristics and their components \citep{burke2013hyperheuristics,stutzle2019automated}. Large language models (LLMs) expand this design space by expressing algorithmic ideas in natural language and generating executable programs \citep{romera2024funsearch,liu2024eoh}. Evaluation and evolutionary search then support repeated selection and revision of these programs using measured performance and execution feedback \citep{vanstein2025llamea}. Subsequent approaches use this experience to provide reflective guidance and organize development trajectories into trees for structured exploration \citep{ye2024reevo,zheng2025mcts}. Reinforcement learning enables the LLM to improve its generation behavior from evaluated candidates \citep{surina2025evotune,huang2026calm} and learn decisions within multi-turn AHD interactions \citep{lv2026ahdagent}. These advances highlight the value of experience in improving heuristic generation.

Algorithm discovery also requires deciding which candidate program to revise and evaluate next. Different revision branches exhibit different patterns of improvement, failure, and stagnation. How the search budget is allocated among these branches, and when exploration stops, therefore directly influence the quality of the algorithms discovered. Historical trajectories record these patterns together with the outcomes of earlier revisions, providing evidence for learning how to make such decisions. We therefore ask: \emph{how can historical search trajectories help us learn better exploration policies?}

World-model research suggests a useful principle: experience can provide an internal environment for improving policies before they guide further real interaction \citep{sutton1990dyna,ha2018worldmodels,hafner2020dreamer,hafner2025dreamerv3}. For algorithm discovery, this motivates turning past searches into an environment where exploration policies can be evaluated before allocating new generation and evaluation requests.

Our key insight is that historical search trees record both what was discovered and how discovery unfolded, providing exactly the structure needed for such an internal environment. We introduce \method{}, which learns search policies by \emph{dreaming} in an experience world constructed from these trees (Figure~\ref{fig:intro-paradigms}c). The relationships among programs, together with their recorded evaluation outcomes, allow candidate search policies to be replayed over these development trajectories. Particularly, a \textit{policy agent} proposes executable search policies that decide which revision branch to extend, when to open a new branch, and when to stop. During replay, each policy sees only the outcomes revealed by its own actions and is evaluated by the quality of the programs it reveals and the number of search steps it uses. These evaluations guide subsequent policy proposals and policy selection. The selected policy then directs a \textit{discovery agent}'s program generation and evaluation, producing new search trajectories that expand the experience world. This cycle turns each completed search into two assets: candidate algorithms and experience for learning more effective exploration policies.

Our contributions are threefold. (1) We formulate historical search trees as an experience world, enabling policies to be compared through recorded outcomes without additional solver generation or evaluation during replay. (2) We develop Dream-COP, which proposes executable policies for budget allocation, and selects candidate policies using a replay objective that balances algorithm quality against the number of search steps. (3) We conduct experiments across five COPs, and the results show that Dream-COP discovers better algorithms with fewer search requests than fixed exploration. Meanwhile, it also achieves superior performance against existing LLM-based AHD methods. Cross-problem experiments further show that exploration policies learned from search experience on one COP remain effective when guiding algorithm discovery on another.

\section{Related Work}

\paragraph{LLM-based algorithm discovery.}

Automated algorithm design builds on configuration methods that optimize parameter settings from measured performance \citep{hutter2011smac,lopez2016irace} and neural approaches that learn solution-construction policies for recurring optimization problems \citep{khalil2017learning,kool2019attention}. 
LLM-based discovery generates executable programs, using evaluated candidates as the basis for proposing new algorithms \citep{liu2023ael,romera2024funsearch,zhu2026simpleevol}. Existing frameworks can jointly develop natural-language heuristic ideas and their implementations \citep{liu2024eoh}, generate complete metaheuristics and solvers \citep{vanstein2025llamea,zhang2026afl}, and extend evaluator-guided code evolution to broader scientific and computational problems \citep{novikov2025alphaevolve}. At the same time, search organization determines which programs receive further development. Reflective methods use performance feedback to guide program revision \citep{ye2024reevo,malik2026pyvrpplus}, while MCTS-AHD organizes heuristic development into a tree to explore promising continuations \citep{zheng2025mcts}. AB-MCTS uses external feedback to choose adaptively between generating new candidates and refining existing ones \citep{inoue2025abmcts}. PathWise organizes heuristic discovery around an entailment graph that records derivation relationships and evaluation history \citep{gungordu2026pathwise}. These methods demonstrate the value of feedback for directing search. Differently, \method{} makes the allocation rule itself an object of program discovery: an LLM proposes executable policies that choose which revision branch to develop and when to stop, and historical replay evaluates these policies before deployment.

\paragraph{Experience-driven search-policy adaptation.}

Experience can change both how candidates are generated and how discovery decisions are made. EvoPH co-evolves prompts and heuristics using feedback from past search \citep{liu2026evoph}. CALM and RFTHGS use evaluated algorithms to improve the LLM generation through reinforcement learning \citep{huang2026calm,zhu2026refining}. AHD Agent extends learning to multi-turn heuristic design, training an LLM to decide between generating heuristics and using tools to obtain evidence in synthesized environments \citep{lv2026ahdagent}. These works adapt prompts, model parameters, or interactive generation behavior. \method{} learns a separate executable controller over the program-search tree, updating budget allocation decisions while keeping LLM weights fixed.

Language agents also adapt through textual feedback, from iterative output refinement to reflections retained across trials \citep{madaan2023selfrefine,shinn2023reflexion}. Agent-Pro optimizes behavioral policies through reflection on trajectories and beliefs \citep{zhang-etal-2024-agent}, while MetaReflection learns instructions from past reflections \citep{gupta-etal-2024-metareflection}. We likewise use accumulated feedback to guide policy proposals. To evaluate a proposed policy, we execute it on the same historical trees and compare the quality of the programs it reveals and the number of search steps it uses.

\paragraph{Policy learning in internal environments.}

Dyna couples real interaction with planning through a learned environment model \citep{sutton1990dyna}. World Models train controllers inside a learned recurrent environment and transfer them to real interaction \citep{ha2018worldmodels}. Dreamer learns behavior from imagined trajectories in latent world models \citep{hafner2020dreamer,hafner2021dreamerv2,hafner2025dreamerv3}. These approaches use internal environments to improve policies before further interaction. \method{} applies this principle to algorithm discovery through an experience world whose transitions and outcomes come directly from recorded program revisions. A policy's actions reveal these records in sequence, allowing alternative allocation and stopping strategies to be compared without new algorithm generation or evaluation during replay. The selected policy then controls subsequent real search, which supplies new trees for learning. This connects experience reuse to an explicit cycle of exploration-policy proposal, replay evaluation, and deployment.

\section{Dream-COP}
\label{sec:method}
\begin{figure*}[t]
  \centering
  \includegraphics[width=\textwidth]{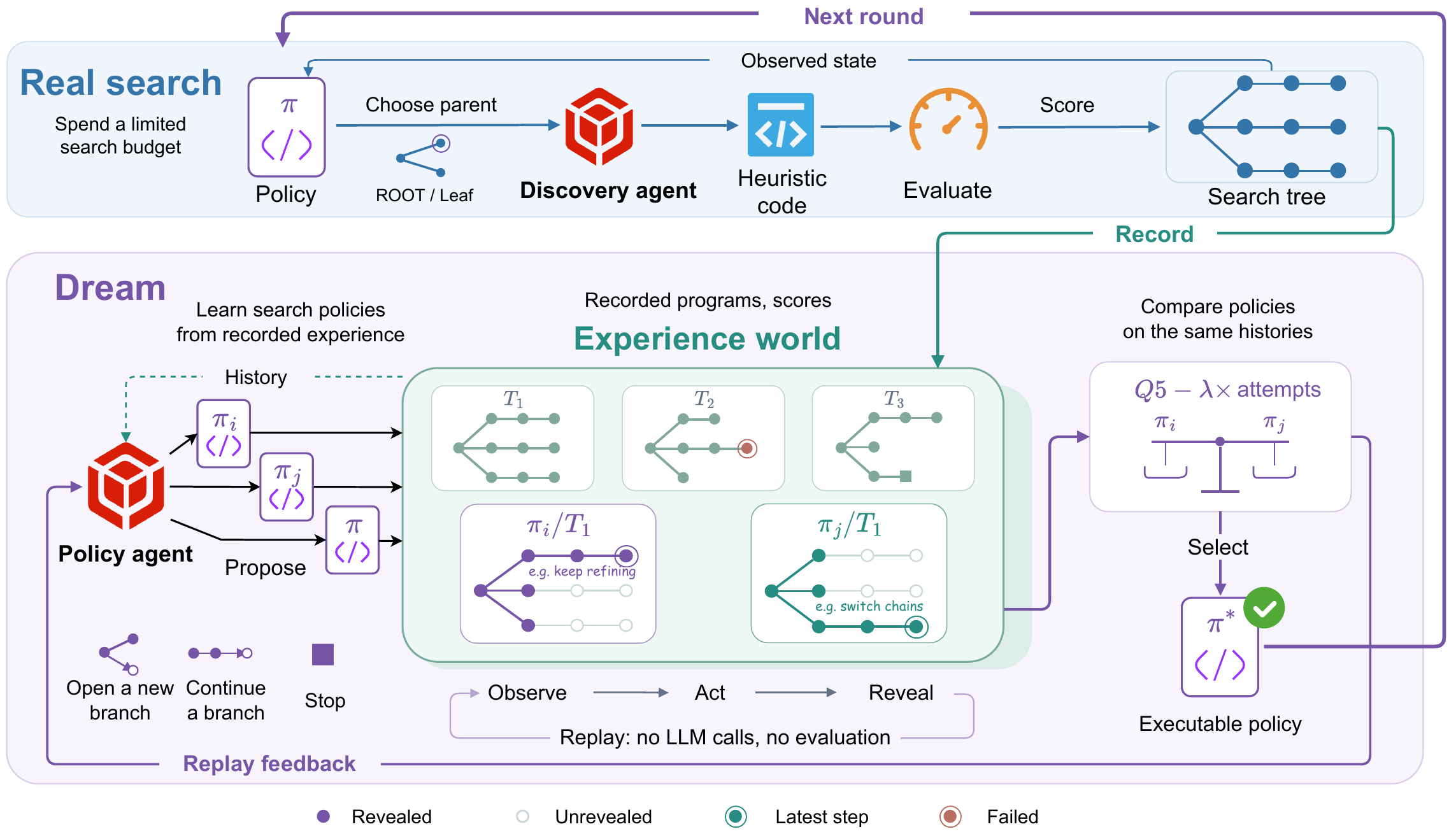}
\caption{\textbf{Learning search policies from recorded experience.} The discovery agent generates and evaluates candidate programs, recording their development trajectories as historical search trees. These trees provide an experience world in which the policy agent compares executable policies. During replay, each policy sees only the outcomes revealed by its own actions; the world supplies recorded transitions without new solver generation or evaluation. Replay feedback guides successive policy proposals, and the selected policy controls the next real search round. The depicted trajectories illustrate alternative search allocations.}
  \label{fig:framework}
  \vspace{-0.5em}
\end{figure*}

\method{} learns search policies by replaying completed search trees (Figure~\ref{fig:framework}). Particularly, a \emph{discovery agent} generates and evaluates solver programs, recording their development as search experience; a \emph{policy agent} uses this experience to learn executable policies that allocate subsequent search budget and decide when to stop. Both agents are instantiated with the same LLM backbone.

\subsection{The Discovery Agent and Search Trees}
Let $p$ denote a solver program and let $q(p;\dev)$ be its quality on a fixed development set, normalized so that larger values are better. The discovery agent generates either a heuristic within a fixed controller or a complete solver. An independent evaluator verifies solutions and computes their quality, providing feedback under either interface.

A search round begins with a feasible root program. At each \emph{search step}, the discovery agent receives the task specification, the selected parent program, the hypotheses associated with its revision branch, and evaluation feedback. It returns a revised program together with a brief description of the change. The evaluator then records the result. Each request consumes one search step, including requests that fail to produce a valid revision.

We record each round as a rooted tree $T=(V,E)$, where only the root may have multiple children, and every other node has at most one child. The tree thus represents alternative branches of program revisions. At the same time, a search policy (i.e., generated by the policy agent) chooses whether to open a new branch, extend an existing branch, or stop.

\subsection{Executable Policies for Budget Allocation}
\label{sec:rounds}
At step $t$, the observation $o_t$ summarizes the revealed tree together with the remaining budget and action history. Then, an executable policy maps this observation to a single action:
\begin{equation}
 a_t=\pi(o_t,\alpha)\in
 \{\texttt{ROOT}\}\cup\mathcal{L}_t\cup\{\texttt{None}\}.
 \label{eq:action}
\end{equation}
Here, $\mathcal{L}_t$ is the set of current leaf identifiers; the policy selects \texttt{ROOT} to open a new branch, $\ell \in \mathcal{L}_t$ to extend the corresponding branch; or \texttt{None} to stop the round. The policy agent proposes policies between rounds; during real search, \dream{} executes the selected policy after observing each completed outcome. Its input $\alpha$ expresses the preference between further exploration and earlier stopping. Appendix~\ref{sec:contract} details the observation fields and presents an example of a learned policy. For comparison, \fixed{} follows a predetermined policy that opens $B$ revision branches and allocates $L$ search steps to each branch. \dream{} uses this policy in its first round. Both methods run for $R$ rounds with a budget of $BL$ search steps per round. Learned policies can redistribute this budget across branch counts and lengths, or stop early.

\subsection{Historical Trees as an Experience World}
\label{sec:world}
The experience world allows a policy to explore the recorded transitions of $T=(V,E)$ in its own order. Let $V_t\subseteq V$ be the nodes revealed after $t$ replay steps and $b_t$ the remaining replay budget. Initially, $V_0$ contains only the root. For a non-stopping action $a_t$ with $b_t>0$, replay updates
\begin{equation}
 \begin{aligned}
  V_{t+1}&=V_t\cup C_T(a_t,V_t),\\
  b_{t+1}&=b_t-1.
 \end{aligned}
 \label{eq:replay-transition}
\end{equation}
Here, $C_T(a_t,V_t)$ contains the next unrevealed child of the root in recorded order, or the recorded child of the selected leaf. It is empty when no continuation remains. Newly revealed nodes expose their stored feasibility and quality outcomes without calling the discovery agent or the evaluator.

During replay, the policy uses the same observation interface as in real search and sees only the outcomes revealed by its actions, including feasibility, solution quality, and execution failures. The same recorded tree can therefore support different search trajectories: a policy may stop after exploring one branch or alternate among branches. These decisions determine which programs are revealed and how many replay steps are used.

Each request to the root or a leaf counts as one \emph{replay step}, including empty requests with $C_T(a_t,V_t)=\varnothing$. Replay ends when the policy returns \texttt{None}, the replay budget is exhausted ($b_t=0$), or all recorded nodes have been revealed ($V_t=V$). Appendix~\ref{sec:contract} specifies replay budgets and eligibility rules for the accumulated history $\hist_r$ after round $r$.

\subsection{Replay Objective}
\label{sec:objective}
A useful search policy should reveal high-quality programs while avoiding unproductive requests. For policy $\pi$ and exploration setting $\alpha$, let $S_{\pi,\alpha,T}$ denote the multiset of development scores of feasible nodes revealed during replay on tree $T$, including the score of the root node. With $m=|S_{\pi,\alpha,T}|$, sort these scores as $s_{(1)}\geq\cdots\geq s_{(m)}$ and define
\begin{equation}
 Q(\pi,\alpha,T)=\frac{1}{k}\sum_{i=1}^{k}s_{(i)},
 \ k=\min(5,m).
 \label{eq:q}
\end{equation}
When development scores are closely clustered, the best score alone provides a narrow view of a policy's search performance. We therefore average the top five feasible scores to capture progress across several promising candidates. Five serves as a practical cutoff that keeps this quality measure focused on high-performing programs while limiting the influence of lower-ranked candidates.

Let $A(\pi,\alpha,T)$ count replay steps, then the replay objective at boundary $r$ is
\begin{align*}
 U_r(\pi,\alpha,T)&=Q(\pi,\alpha,T)-\lambda A(\pi,\alpha,T),
 \\
 J_r(\pi,\alpha)&=\frac{1}{|\hist_r^\star|}
  \sum_{T\in\hist_r^\star}U_r(\pi,\alpha,T),
\end{align*}
where $\hist_r^\star$ is the nonempty set of eligible trees available at this boundary, weighted equally. The fixed coefficient $\lambda$ penalizes each replay step in units of normalized development quality. Thus, $\alpha$ controls exploration preference, while $\lambda$ determines the trade-off between program quality and search cost in the replay objective $J_r(\pi,\alpha)$.

\subsection{The Policy Agent: Learning from Replay}
\label{sec:learning}
Let $c_r=(\pi_r,\alpha_r)$ be the pair deployed in round $r$. Starting with $\hist_0=\varnothing$, the first round uses $\pi_1=\fixed{}$, which ignores its $\alpha$ input. Each round generates a search tree and expands the history:
\begin{equation}
 \begin{aligned}
  T_r&=\operatorname{Search}(\pi_r,\alpha_r;BL),\\
  \hist_r&=\hist_{r-1}\cup\{T_r\},
 \end{aligned}
 \label{eq:history-update}
\end{equation}
where $\operatorname{Search}$ executes the discovery agent under the deployed policy with at most $BL$ search steps.

Before the next round ($r<R$), policy learning runs whenever the expanded history contains eligible trees, i.e., $\hist_r^\star\neq\varnothing$.
The policy agent proposes executable policies through four sequential LLM calls, using the current policy and historical evidence. After each call, replay evaluates the proposals; rewards, quality, step counts, and failures inform the next call. This feedback guides refinement of stopping, branching, allocation, and revisiting decisions. Appendix~\ref{sec:policy-prompt} provides the prompt template.

For a policy--parameter pair $c=(\pi,\alpha)$, write $J_r(c)=J_r(\pi,\alpha)$. Let $\mathcal{C}_r$ contain the valid pairs generated at this boundary, then the pair with the highest mean replay reward is deployed as:
\begin{equation}
 c_{r+1}=
 \begin{cases}
  \arg\max_{c\in\mathcal{C}_r}J_r(c),
       &\mathcal{C}_r\neq\varnothing,\\
  c_r,&\mathcal{C}_r=\varnothing.
 \end{cases}
 \label{eq:update}
\end{equation}
Policy learning leaves LLM weights unchanged. The selected policy remains fixed throughout the next real search round, which supplies new experience. Appendix~\ref{sec:contract} provides execution and selection details and the complete procedure (Algorithm~\ref{alg:dream}).

\section{Experiments}
\label{sec:experiments}
Our experiments address three questions: (1) Can \method{} discover better algorithms with fewer search requests? (2) Does historical replay improve the effectiveness of learned search policies? (3) How well do these policies transfer across different COPs?

We first evaluate \method{} on four tasks in which the discovery agent evolves heuristic components within fixed controllers: constructive variants of the traveling salesman problem (TSP) and knapsack problem (KP), online bin packing (OBPP), and flow-shop scheduling (FSSP). We further use offline two-dimensional bin packing (2D-BPP) to evaluate complete solver generation and to analyze policy replay and the trade-off between solution quality and search cost. Moreover, we evaluate cross-problem policy transfer among TSP, OBPP, and FSSP. Complete \dream{}-discovered programs are provided in Appendix~\ref{sec:discovered-programs}.

\begin{table*}[!t]
\centering\small
\setlength{\tabcolsep}{3pt}
\resizebox{0.98\textwidth}{!}{%
\begin{tabular}{@{}llrrrrrrrrrr@{}}
\toprule
\multirow{2}{*}{LLM Model} & \multirow{2}{*}{Method} &
  \multicolumn{2}{c}{$N=50$} & \multicolumn{2}{c}{$N=100$} &
  \multicolumn{2}{c}{$N=200$} & \multicolumn{2}{c}{$N=500$} &
  \multicolumn{2}{c}{TSPLIB} \\
 & & Obj.$\downarrow$ & Gap (\%) & Obj.$\downarrow$ & Gap (\%) & Obj.$\downarrow$ &
  Gap (\%) & Obj.$\downarrow$ & Gap (\%) & Obj.$\downarrow$ & Gap (\%) \\
\midrule
\multirow{3}{*}{} & LKH ref. & 5.675 & -- & 7.768 & -- & 10.708 & -- &
  16.531 & -- & 46392.7 &
  -- \\
 & Greedy & 6.959 & 22.622 & 9.706 & 24.936 & 13.461 & 25.711 & 20.813 & 25.903 &
  57055.2 & 24.332 \\
 & POMO & \textbf{5.697} & \textbf{0.383} & \textbf{8.001} & \textbf{2.999} & 12.897 &
  20.447 & 24.156 & 46.135 & 60654.7 & 23.923 \\
\midrule
\multirow{5}{*}{GPT-4o-mini} & EoH & 6.629 & 16.822 & 9.252 & 19.114 & 12.939 &
  20.842 & 20.202 & 22.209 & 55974.5 &
  21.060 \\
 & ReEvo & 6.959 & 22.622 & 9.706 & 24.936 & 13.461 & 25.711 & 20.813 & 25.903 &
  57055.2 & 24.332 \\
 & MCTS-AHD & 6.578 & 15.906 & 9.231 & 18.832 & 12.974 & 21.166 & 20.258 & 22.549 &
  55759.7 & 19.306 \\
 & \fixed{} & 6.959 & 22.622 & 9.706 & 24.936 & 13.461 & 25.711 & 20.813 & 25.903 &
  57055.2 & 24.332 \\
 & \method{} & \cellcolor{gray!18}6.490 & \cellcolor{gray!18}14.379 &
  \cellcolor{gray!18}9.006 & \cellcolor{gray!18}15.952 & \cellcolor{gray!18}12.587 &
  \cellcolor{gray!18}17.557 & \cellcolor{gray!18}19.696 & \cellcolor{gray!18}19.149 &
  \cellcolor{gray!18}54416.3 & \cellcolor{gray!18}17.390 \\
\midrule
\multirow{5}{*}{GPT-6-Luna} & EoH & 6.630 & 16.819 & 9.241 & 18.955 & 12.882 &
  20.305 & 20.019 & 21.099 & 56590.2 & 20.658 \\
 & ReEvo & 6.566 & 15.682 & 9.204 & 18.493 & 12.881 & 20.292 & 20.075 & 21.440 &
  55454.1 & 20.782 \\
 & MCTS-AHD & 6.582 & 15.959 & 9.161 & 17.929 & 12.758 & 19.143 & 19.792 & 19.728 &
  55893.7 & 20.226 \\
 & \fixed{} & 6.543 & 15.275 & 9.123 & 17.432 & 12.688 & 18.495 & 19.734 & 19.378 &
  54476.4 & 18.355 \\
 & \method{} & \cellcolor{gray!18}6.214 & \cellcolor{gray!18}9.489 &
  \cellcolor{gray!18}8.647 & \cellcolor{gray!18}11.316 &
  \cellcolor{gray!18}\textbf{12.031} & \cellcolor{gray!18}\textbf{12.361} &
  \cellcolor{gray!18}\textbf{18.714} & \cellcolor{gray!18}\textbf{13.209} &
  \cellcolor{gray!18}\textbf{51666.6} & \cellcolor{gray!18}\textbf{11.984} \\
\bottomrule
\end{tabular}
} \caption{Constructive TSP: mean tour length and gap (\%) to LKH, which matches the official optima on all 15 TSPLIB instances. We also include Greedy and POMO \citep{kwon2020pomo} as baselines. LLM-based results are averaged over 3 runs. Bold: best non-reference result; shading: best LLM result within each model, including ties.}
\label{tab:constructive}
\end{table*}

\begin{table*}[!t]
\centering\small
\setlength{\tabcolsep}{3pt}
\resizebox{0.98\textwidth}{!}{%
\begin{tabular}{@{}llrrrrrrrr@{}}
\toprule
\multirow{2}{*}{LLM Model} & \multirow{2}{*}{Method} &
  \multicolumn{2}{c}{$N=50$, $W=12.5$} & \multicolumn{2}{c}{$N=100$, $W=25$} &
  \multicolumn{2}{c}{$N=200$, $W=25$} & \multicolumn{2}{c}{$N=500$, $W=25$} \\
 & & Obj.$\uparrow$ & Gap (\%) & Obj.$\uparrow$ & Gap (\%) &
  Obj.$\uparrow$ & Gap (\%) & Obj.$\uparrow$ & Gap (\%) \\
\midrule
\multirow{3}{*}{} & OR-Tools opt. & 20.037 & -- & 40.272 & -- & 57.447 & -- & 90.963 & -- \\
 & Greedy & 19.985 & 0.261 & 40.225 & 0.118 & 57.395 & 0.090 & 90.919 & 0.048 \\
 & POMO & 19.510 & 2.548 & 39.431 & 2.034 & 57.209 & 0.417 & 85.414 & 6.096 \\
\midrule
\multirow{3}{*}{GPT-6-Luna} & MCTS-AHD &
  \cellcolor{gray!18}\textbf{20.037} & \cellcolor{gray!18}\textbf{0.000} &
  \cellcolor{gray!18}\textbf{40.272} & \cellcolor{gray!18}\textbf{0.000} &
  \cellcolor{gray!18}\textbf{57.447} & \cellcolor{gray!18}\textbf{0.000} & 90.945 & 0.020 \\
 & \fixed{} & 20.033 & 0.019 & 40.264 & 0.020 & 57.436 & 0.020 & 90.954 & 0.010 \\
 & \method{} &
  \cellcolor{gray!18}\textbf{20.037} & \cellcolor{gray!18}\textbf{0.000} &
  \cellcolor{gray!18}\textbf{40.272} & \cellcolor{gray!18}\textbf{0.000} &
  \cellcolor{gray!18}\textbf{57.447} & \cellcolor{gray!18}\textbf{0.000} &
  \cellcolor{gray!18}\textbf{90.963} & \cellcolor{gray!18}\textbf{0.000} \\
\bottomrule
\end{tabular}
} \caption{Constructive KP: mean objective value and gap (\%) to the OR-Tools optimum by item count $N$ and capacity $W$, averaging 1,000 test instances per size. LLM-based methods use GPT-6-Luna. Bold: best non-reference result; shading: best LLM result, including ties.}
\label{tab:kp}
\end{table*}

\subsection{Evaluation Settings}
\label{sec:selection}

\paragraph{Tasks and instances.}
For constructive TSP and KP, we aim to evolve next-city and next-item selectors within fixed construction procedures to minimize tour length and maximize total item value, respectively. Each task is evaluated on 1,000 test instances at each size $N\in\{50,100,200,500\}$. TSP additionally includes 15 TSPLIB instances, totaling 4,015 test instances. For OBPP, we aim to evolve a bin-scoring rule to minimize bin count and evaluate it on 30 test instances spanning 1,000--10,000 items and capacities 100 and 500. For FSSP, we aim to evolve a perturbation function within a fixed local-search controller to minimize makespan and evaluate it on 110 Taillard test instances covering 11 sizes \citep{taillard1993benchmarks}. For 2D-BPP, we aim to evolve complete solvers to minimize bin count and evaluate them on 500 BWMV test instances with 20--100 rectangles per instance \citep{martello1998exact}.

Following MCTS-AHD \citep{zheng2025mcts}, we use its released development sets of 64 TSP instances with 50 cities, 64 KP instances with 100 items, and four OBPP instances. For FSSP, we use the 64 development instances with 50 jobs released in EoH \citep{liu2024eoh}. For 2D-BPP, we generate 60 instances following the BWMV class distributions \citep{martello1998exact}.  These development sets are disjoint from the corresponding test sets. Task specifications are detailed in Appendix~\ref{sec:task-details}.

\paragraph{Methods and setting.}
We compare with applicable implementations of FunSearch, EoH, ReEvo, and MCTS-AHD \citep{romera2024funsearch,liu2024eoh,ye2024reevo,zheng2025mcts}. Both \fixed{} and \dream{} use the same discovery agent and run for $R=10$ rounds. \fixed{} opens $B=6$ branches per round with a maximum depth of $L=4$, whereas \dream{} adaptively allocates the same budget of $BL=24$ search steps per round and may stop early. Both therefore allow at most 240 search steps overall; the other baselines receive a budget of 240 candidates, including initialization. Invalid generations and failed evaluations count toward these budgets. Search steps measure algorithm-discovery effort, whereas token costs also account for policy learning. The main backbone is GPT-6-Luna; TSP and FSSP also use GPT-4o-mini, and 2D-BPP also uses GPT-6-Sol. In the GPT-6-Luna experiments on TSP and 2D-BPP, \fixed{} and \dream{} share the initial solver within each comparison. For each run, we freeze the solver with the highest development score before testing it directly on all test instances. All reported evaluations are complete, with zero failed instances. Appendix~\ref{sec:run-settings} provides detailed experimental configurations.

\Needspace{5\baselineskip}
\subsection{Performance across Problems and Sizes}

\paragraph{Constructive TSP and KP.}
Table~\ref{tab:constructive} shows that \dream{} achieves the lowest mean tour length among the compared LLM-based AHD methods across all four synthetic sizes and TSPLIB, using either LLM backbone, i.e., GPT-6-Luna or GPT-4o-mini. Here, with GPT-6-Luna, it reduces the overall mean gap to LKH from \fixed{}'s 17.65\% to 11.60\% while using 13.75\% fewer search requests. 

\begin{wrapfigure}{r}{0.48\textwidth}
\centering
\includegraphics[width=\linewidth]{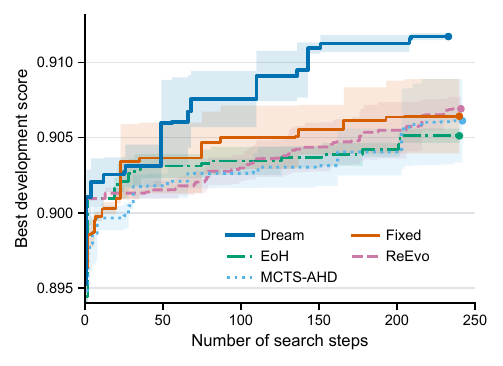}
\caption{TSP development score using GPT-6-Luna, averaged over three independent runs per method.}
\label{fig:tsp-evolution}
\end{wrapfigure}

These gains show that learned search policies discover constructive programs that generalize beyond the development instances. At $N=500$, \dream{} also reduces the mean solve time per instance from 137 seconds with \fixed{} to 80 seconds, improving both tour quality and execution efficiency (Appendix~\ref{sec:tsp-times}). Figure~\ref{fig:tsp-evolution} shows the mean best-so-far development score across three runs, with shading indicating $\pm1$ standard error. \dream{} reaches a higher final score than \fixed{} while using fewer search requests. It also surpasses the compared LLM-based baselines, showing that more selective use of the search budget can improve the quality of the discovered programs.

Table~\ref{tab:kp} reports the constructive KP results across four item counts. \dream{} matches the OR-Tools optimal values to the displayed precision at all four sizes, whereas none of the compared baselines consistently achieves a zero gap. At the largest size, $N=500$, MCTS-AHD and \fixed{} retain gaps of 0.020\% and 0.010\%, respectively. \dream{} attains its higher solution quality while using 17.5\% fewer search requests than \fixed{}, demonstrating more effective use of the discovery budget.

\begin{figure}[tbp]
\centering
\begin{minipage}[t]{0.48\textwidth}
\vspace{0pt}
\centering
\includegraphics[width=\linewidth]{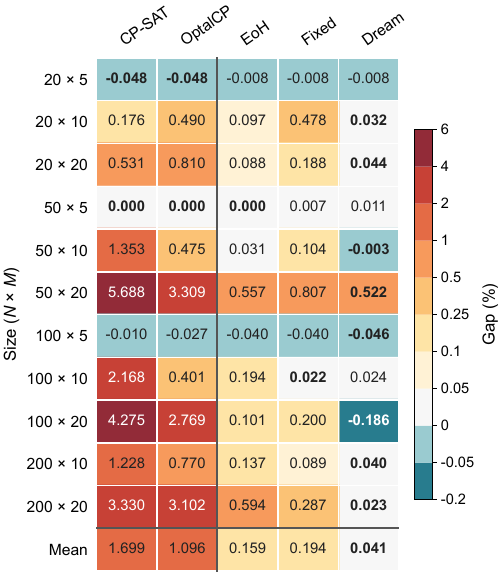}
\captionof{figure}{FSSP mean gap (\%) to the Taillard reference upper bound by size (lower is better). LLM-based methods use GPT-6-Luna. Bold marks the best method per row.}
\label{fig:fssp-gap}
\end{minipage}\hfill
\begin{minipage}[t]{0.48\textwidth}
\vspace{0pt}
\centering\footnotesize
\setlength{\tabcolsep}{2pt}
\resizebox{\linewidth}{!}{%
\begin{tabular}{@{}lrrrrrr@{}}
\toprule
$N$ & FunSearch & EoH & ReEvo & MCTS & \fixed{} & \dream{} \\
\midrule
1k & 4.771 & 4.771 & 4.522 & \cellcolor{gray!18}3.926 & 4.371 & 4.477 \\
5k & 4.307 & 4.307 & 3.743 & 3.357 & 3.585 & \cellcolor{gray!18}1.089 \\
10k & 4.049 & 4.049 & 3.730 & 3.291 & 3.501 & \cellcolor{gray!18}0.459 \\
\bottomrule
\end{tabular}
} \captionof{table}{OBPP with GPT-6-Luna: mean gap (\%) to the volume lower bound at capacity $C=100$, by item count (lower is better). Shading marks the best method at each size.}
\label{tab:obpp}

\par\vspace{1.6em}
\centering\small
\setlength{\tabcolsep}{3pt}
\resizebox{\linewidth}{!}{%
\begin{tabular}{@{}lrrrrr@{}}
\toprule
\multirow{2}{*}{$N$} & BRKGA & \multicolumn{2}{c}{GPT-6-Luna} &
  \multicolumn{2}{c}{GPT-6-Sol} \\
\cmidrule(lr){3-4}\cmidrule(lr){5-6}
 & Total & \fixed{} & \dream{} & \fixed{} & \dream{} \\
\midrule
20 & 515 & +0.3 & \cellcolor{gray!18}0.0 & \cellcolor{gray!18}0.0 &
  \cellcolor{gray!18}0.0 \\
40 & 977 & +2.0 & \cellcolor{gray!18}+1.0 & \cellcolor{gray!18}+1.0 & \cellcolor{gray!18}+1.0 \\
60 & 1446 & +6.3 & \cellcolor{gray!18}+4.7 & +2.7 & \cellcolor{gray!18}+1.3 \\
80 & 1964 & +5.0 & \cellcolor{gray!18}+3.3 & +2.3 & \cellcolor{gray!18}+0.3 \\
100 & 2332 & +12.0 & \cellcolor{gray!18}+8.7 & +5.3 & \cellcolor{gray!18}+3.0 \\
All & 7234 & +25.7 & \cellcolor{gray!18}+17.7 & +11.3 & \cellcolor{gray!18}+5.7 \\
\bottomrule
\end{tabular}
} \captionof{table}{2D-BPP: BRKGA total bins and excess bins ($\Delta$) for \fixed{}/\dream{}, averaged over three runs. Each size has 100 instances (500 total). Shading marks the smaller excess within each model, including ties.}
\label{tab:fullsolver}
\end{minipage}
\end{figure}

\paragraph{Online packing and flow-shop scheduling.}
Table~\ref{tab:obpp} reports OBPP gaps at capacity $C=100$, with each cell averaging five instances. Table~\ref{tab:obpp-details} in the Appendix includes both $C=100$ and $C=500$. Averaged over all 30 test instances across both capacities, \dream{} reduces the gap to the volume lower bound from \fixed{}'s 2.112\% to 1.199\%, using 42.5\% fewer search steps. The improvement is concentrated in longer item sequences at capacity 100, where \dream{} outperforms all included baselines. Since every method uses the same packing controller, this gain reflects a more effective bin-scoring rule discovered through adaptive search.

On FSSP, \dream{} achieves the lowest aggregate makespan among the LLM-based methods with both LLM backbones, with improvements over \fixed{} and EoH extending to the largest job counts. With GPT-6-Luna, its mean gap to the upper bound is 0.0412\%. Across the 110 instances, \dream{} records 59 wins, 40 ties, and 11 losses against \fixed{}. Excluding ties, a two-sided exact sign test gives $p=4.47\times10^{-9}$. This extends the benefit of learned search from constructive decisions to perturbations within iterative local search. Figure~\ref{fig:fssp-gap} further shows that \dream{}'s 60-second GLS protocol achieves a lower mean gap than CP-SAT and OptalCP under their 300-second limits. The LLM-based results in the heatmap use GPT-6-Luna, where each entry for a given problem size averages ten instances, while the overall mean averages all 110 instances. Appendix~\ref{sec:fssp-makespan} provides the complete makespan/time results (Table~\ref{tab:fssp-makespan}) and settings.

\subsection{Designing Complete 2D-BPP Solvers}
\label{sec:fullsolver}
The preceding tasks evolve components within fixed controllers. For 2D-BPP, we instead ask the discovery agent to generate complete solvers that assign rectangles to bins and determine their positions without rotation. Each solver implements its own construction and improvement logic within a given time budget. An independent geometric checker verifies the resulting packing layouts.

We evaluate candidate solvers on 60 development instances using mean squared bin utilization as the score. The selected solvers are then tested on 500 BWMV instances with a 60-second budget per instance. Table~\ref{tab:fullsolver} reports their excess bin counts relative to the published BRKGA reference \citep{goncalves2013brkga}. All selected solvers produce feasible layouts on every instance, and \dream{} uses fewer bins than \fixed{} with both settings. With GPT-6-Luna, it saves eight bins over the 500-instance test set, with most of the improvement occurring on larger instances. These results show that learned search policies remain effective when the discovery agent must coordinate construction, placement, and improvement within a complete solver.

\Needspace{6\baselineskip}
\subsection{Experimental Analysis}
\label{sec:2dbpp-allocation}

We examine the two terms of the replay objective by comparing quality measures across all five COPs and varying the cost penalty on 2D-BPP. We then use 2D-BPP to assess the contribution of policy replay and illustrate how learned policies allocate the search budget.

\paragraph{Replay quality measure.}
The replay quality term $Q$ in Eq.~\eqref{eq:q} averages the top five feasible development scores. We denote it by $Q_5$ and compare it with $Q_1=\max S_{\pi,\alpha,T}$, keeping the request penalty unchanged. Whereas $Q_1$ increases only when a new best score is revealed, $Q_5$ also captures improvements in the overall quality of the top-five candidates. We examine whether this richer signal helps exploration through periods without a new best. Both objectives use GPT-6-Luna with the same root programs and budgets for discovery and policy learning, with one completed run per objective and problem.

\noindent
\begin{minipage}[t]{0.48\textwidth}
\vspace{0pt}
We count a delayed improvement when a new best development score is reached within a round after at least three consecutive requests without a new best. To account for varying search effort, we report these events per 100 discovery requests across the learned-policy rounds, excluding the initial round. Failed requests count toward both the stagnation interval and the request total. Figure~\ref{fig:replay-quality-delayed} shows higher delayed-improvement rates with $Q_5$ on all five problems, suggesting that the quality of several promising candidates provides useful guidance during stagnation.
\end{minipage}\hfill%
\begin{minipage}[t]{0.48\textwidth}
\vspace{-1em}
\centering
\includegraphics[width=\linewidth]{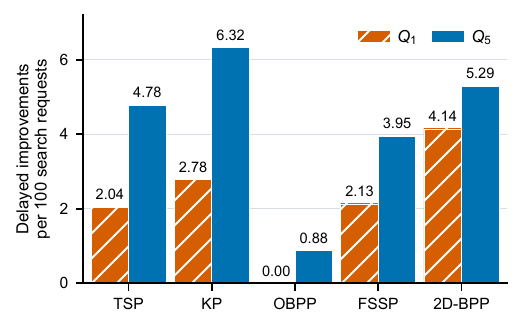}
\vspace{-2em}
\captionof{figure}{Delayed improvements per 100 discovery requests under $Q_1$ and $Q_5$ with GPT-6-Luna.}
\label{fig:replay-quality-delayed}
\end{minipage}
\par\medskip

The selected solvers also achieve higher development scores and better test performance with $Q_5$ on TSP, FSSP, and 2D-BPP, with identical KP results (see Appendix~\ref{sec:replay-quality-ablation}). Together, these results support using multiple promising candidates to assess replay quality and guide continued exploration.

\begin{wrapfigure}{r}{0.48\textwidth}
\centering
\vspace{-0.5em}
\includegraphics[width=0.9\linewidth]{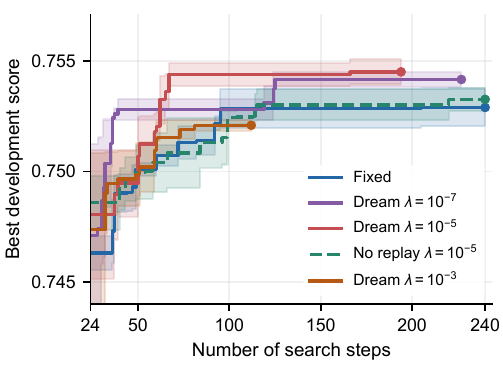}
\vspace{-1em}
\caption{2D-BPP search progress under different cost penalties and with or without policy replay, using GPT-6-Luna. Curves show the best-so-far development score, averaged over three runs per setting.}
\label{fig:2dbpp-generation}
\end{wrapfigure}

\paragraph{Cost penalty.}
We next examine the cost term of the replay objective by varying $\lambda$ on 2D-BPP. Figure~\ref{fig:2dbpp-generation} compares search progress under different penalties and with or without policy replay, averaging three runs per setting. \dream{} with $\lambda=10^{-7}$ and $10^{-5}$ reaches higher mean development scores than \fixed{} with fewer search steps, whereas $\lambda=10^{-3}$ favors shorter searches. Increasing $\lambda$ from $10^{-5}$ to $10^{-3}$ reduces mean search steps by 44.1\%, while the final mean development score changes from 0.75451 to 0.75209. At $\lambda=10^{-3}$, \dream{} uses 106 search steps on average, compared with 240 for \fixed{}, and still achieves a lower mean test bin count (14.51333 versus 14.51933). We use $\lambda=10^{-5}$ as the default to balance quality and cost. Appendix~\ref{sec:2dbpp-cohort} provides the full sensitivity results in Table~\ref{tab:2dbpp-sensitivity} and round-level statistics in Figure~\ref{fig:2dbpp-attempts}.

The reduction in search effort also yields net token savings. Including policy learning, \dream{} at $\lambda=10^{-3}$ reduces mean total recorded tokens by 23.8\% relative to \fixed{}, as the savings in solver discovery exceed the cost of policy learning. Appendix~\ref{sec:2dbpp-cohort} reports the token breakdown.

\dreamwrapclear
\paragraph{Policy-replay ablation.}
Using the default cost penalty, we next examine the contribution of replay-based policy evaluation and selection. We compare \dream{} with \emph{No replay}, which uses the same historical-evidence interface and algorithm-discovery budget. At each learning boundary, No replay makes one LLM call to generate a single policy and deploys it after syntax and interface checks, without replay-based evaluation or selection. In Figure~\ref{fig:2dbpp-generation}, No replay remains close to \fixed{}, whereas \dream{} reaches a higher development score with fewer search steps. Comparing three-run mean bin counts on each of the 500 test instances gives \dream{} 15 wins, 483 ties, and 2 losses against No replay. Excluding ties, a two-sided exact sign test gives $p=0.00235$. These results support the replay-based policy-learning loop, which evaluates policies in the experience world before deployment.

\begin{figure*}[!htbp]
\centering
\vspace{0.5em}
\includegraphics[width=\textwidth]{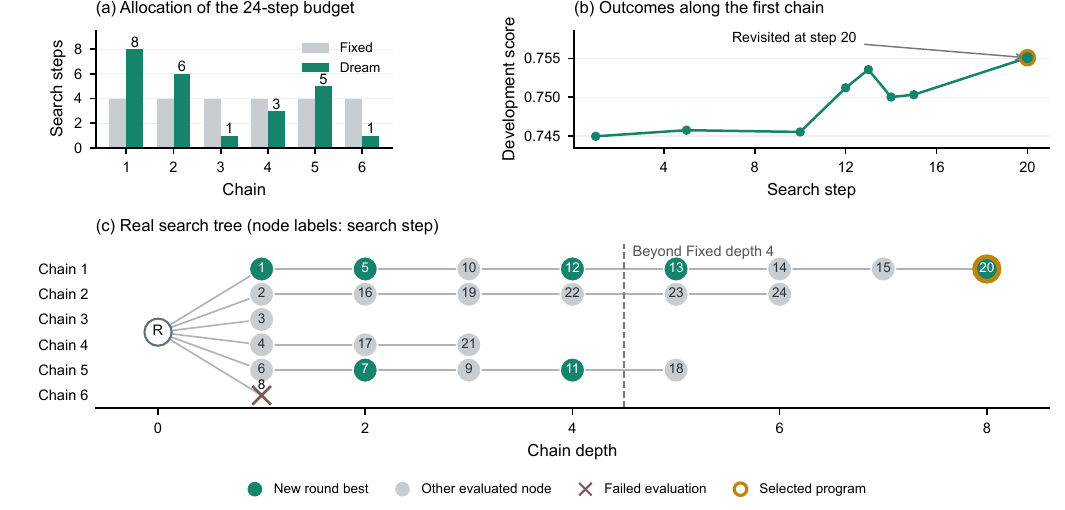}
\vspace{-1.5em}
\caption{2D-BPP search round 9 with GPT-6-Luna. (a) Allocation of the 24-step budget across branches, compared with four steps per branch under \fixed{}. (b) Development scores along the first branch. (c) Complete search tree, with nodes labeled by search step. The highlighted program has the highest development score in the run and is selected for testing.}
\label{fig:2dbpp-policy-trajectory}
\end{figure*}

\paragraph{Learned budget allocation in real search.}
To illustrate how these policies guide individual search decisions, Figure~\ref{fig:2dbpp-policy-trajectory} follows \dream{} within a round's 24-step budget. It allocates different numbers of steps to the six branches, extending promising ones beyond \fixed{}'s four-step limit. After two lower-scoring revisions on the first branch, it explores other branches before returning at step 20. This revisit produces the program with the highest development score in the run, at depth eight. The trajectory shows how learned policies combine deeper refinement with exploration of alternatives under the same total budget.

\subsection{Cross-Problem Policy Transfer}
\label{sec:cross-problem}
We further examine whether learned search policies transfer effectively across COPs. We consider three directions: TSP $\rightarrow$ OBPP, OBPP $\rightarrow$ FSSP, and FSSP $\rightarrow$ TSP. For TSP, we report mean tour length with $N\in\{50,100,200\}$. For each transfer, we deploy the final search policy learned in a \dream{} run on the source problem, keeping its code fixed throughout target search. 

All target runs use GPT-6-Luna, $B=6$, $L=4$, and $R=10$. For each transfer run, we freeze the program with the highest development score before testing. Table~\ref{tab:cross-problem} compares transferred policies with the \fixed{} and \dream{} results. Transferred policies improve test quality over \fixed{} in all three directions while using 25.0--64.2\% fewer search steps. On OBPP, transfer also surpasses \dream{}, reducing the mean gap from 1.1988\% to 1.0637\%. A plausible explanation is that transfer benefits from an exploration strategy already learned on the source problem, which guides OBPP discovery from the first round, whereas \dream{} initially collects target-problem experience before learning its policy online. These results demonstrate that learned exploration policies provide reusable guidance for algorithm discovery across different COPs.

\begin{table}[!htbp]
\centering\small
\setlength{\tabcolsep}{6pt}
\vspace{-1em}
\begin{tabular}{@{}llrrr@{}}
\toprule
Transfer & Method &
Dev. score$\uparrow$ &
Test metric$\downarrow$ &
Search steps$\downarrow$ \\
\midrule
\multirow{3}{*}{TSP $\rightarrow$ OBPP}
 & \fixed{} & 0.98142 & 2.1124 & 240 \\
 & \dream{} & 0.98409 & 1.1988 & 138 \\
 & Transferred \dream{}
 & \textbf{0.98710} & \textbf{1.0637} & \textbf{86} \\
\midrule
\multirow{3}{*}{OBPP $\rightarrow$ FSSP}
 & \fixed{} & 0.54973 & 0.1940 & 240 \\
 & \dream{} & \textbf{0.54990} & \textbf{0.0412} & 176 \\
 & Transferred \dream{}
 & 0.54981 & 0.0872 & \textbf{174} \\
\midrule
\multirow{3}{*}{FSSP $\rightarrow$ TSP}
 & \fixed{} & 0.90642 & 9.4513 & 240 \\
 & \dream{} & \textbf{0.91173} & \textbf{8.9643} & 207 \\
 & Transferred \dream{}
 & 0.91114 & 8.9863 & \textbf{180} \\
\bottomrule
\end{tabular}
\caption{Cross-problem policy transfer with GPT-6-Luna, using one paired run per direction. Test metric reports mean tour length for TSP and mean gap (\%) for OBPP and FSSP. Dev. score reports the highest development score in each run. Search steps count algorithm-discovery requests, including failures, with a maximum of 240 per run. Bold marks the better result within each comparison.}
\label{tab:cross-problem}
\end{table}

\section{Conclusion}
We introduced \method{}, which learns search policies through historical replay to guide algorithm discovery for combinatorial optimization problems. Recorded transitions and outcomes enable executable search policies to be evaluated and refined through replay, without additional solver generation or evaluation. The learned policies then guide budget allocation in subsequent search rounds. Experiments across five COPs demonstrate that this approach discovers better algorithms with fewer search requests than fixed exploration. These gains hold for the generation of both heuristic components and complete solvers. The replay ablation further supports evaluating candidate policies in the experience world before deployment. Cross-problem experiments further show that learned search policies remain effective beyond the problems on which they were learned. Together, these results demonstrate that historical search experience can be used to learn policies that guide more effective algorithm discovery.


\ifshowbiburls
\bibliographystyle{iclr2026_conference}
\else
\bibliographystyle{iclr2026_conference_nourl}
\fi
\bibliography{references}
\clearpage
\appendix

\section{Algorithm and Agent Details}
\label{sec:contract}
Algorithm~\ref{alg:dream} summarizes how solver discovery and policy learning alternate across rounds. Each completed search round contributes a tree to the accumulated history, providing experience for evaluating policy proposals through replay and selecting a policy for subsequent search. The following sections detail the policy interface and observations, history eligibility and replay rules, and the policy-selection procedure. We also provide the core prompts for the discovery and policy agents, followed by an executable policy example that illustrates how the learned decision rules operate.

\begin{algorithm}[!htbp]
\caption{\method{}: real search and policy learning}
\label{alg:dream}
\begin{algorithmic}[1]
\Require Task, reference program, $R,B,L$, penalty $\lambda$
\State $\pi\gets\fixed{}$; $\alpha\gets\texttt{None}$; $\hist\gets\varnothing$
\For{$r=1,\ldots,R$}
 \State Initialize root $v_0^{(r)}$ under the task contract
 \State $T_r\gets\operatorname{Search}(v_0^{(r)},\pi,\alpha,BL)$
 \State $\hist\gets\hist\cup\{T_r\}$
 \State $\hist^\star\gets\operatorname{Eligible}(\hist)$
 \If{$r<R$ and $\hist^\star\neq\varnothing$}
  \State $\mathcal{C}\gets\varnothing$
  \For{$j=1,\ldots,4$}
   \State Propose policies using accumulated feedback
   \State Replay each at five $\alpha$ values on $\hist^\star$
   \State Add valid pairs to $\mathcal{C}$; summarize outcomes
  \EndFor
  \State Update $(\pi,\alpha)$ by Eq.~\eqref{eq:update}
 \EndIf
\EndFor
\State Select and freeze the solver with the highest development score
\State Evaluate the frozen solver directly on the test set
\end{algorithmic}
\end{algorithm}

\paragraph{Policy interface.}
A policy is implemented as \texttt{policy(observation, alpha)}. At each step, it returns \texttt{ROOT} to start a new branch, a current leaf to extend an existing branch, or \texttt{None} to stop the current round. \dream{} executes one search request at a time.

The observation contains the search settings $B$, $L$, and $R$, together with the total, used, and remaining budget. For each node in the revealed search tree, it records the parent, program and development score. It also reports the branch count and identifies the root and current leaves as the points at which the policy can start a new branch or extend an existing one, respectively. Unselected branches remain available for subsequent exploration, even after a failed search step.

To compare branches, the observation provides the global best score and top-five mean, per-branch best and current scores, and quality gaps. Branch depth, recent gains, stagnation, and the number of steps since a branch's last request help the policy decide whether to continue or revisit it. Feasibility, execution status, failures, and empty-request counts distinguish evaluated outcomes from missing continuations. The action history records previous choices and revealed outcomes, along with the budget and quality after each step, allowing the policy to reconstruct its past decisions within the current tree.

A program that produces an infeasible solution receives no numerical quality score. An interrupted request still consumes budget, but is not classified as a solver failure. During replay, an empty request indicates that no recorded continuation is available, whereas a request reaching an invalid program reveals the recorded failure. Both consume one replay step, and the request status is included in the next observation. All observation fields use only information revealed so far, excluding future scores and counts of unrevealed descendants. Completed trees are provided separately to the policy agent for policy learning.

Each policy decision runs in a fresh restricted scope, so the policy must reconstruct its earlier decisions from the supplied action history. Execution permits only approved operations, with no access to files or networks, and is limited to one second and 256 MiB of memory per decision.

\paragraph{History and replay budgets.}
The history $\hist_r$ contains completed, replayable search trees. A tree is eligible for policy learning once it contains sufficiently many recorded search steps:
\begin{equation}
 \hist_r^\star=
 \left\{T\in\hist_r:\ |V(T)|-1\geq
 \left\lceil\frac{BL}{5}\right\rceil\right\}.
 \label{eq:eligible-history}
\end{equation}
Here, $V(T)$ is the node set of $T$. The count excludes the root and includes failed search steps; eligibility does not require an improvement in solution quality. With $B=6$ and $L=4$, a tree becomes eligible after five recorded search steps.
Before each subsequent round ($r<R$), policy learning uses all eligible trees accumulated so far. Learning proceeds whenever this set is nonempty, even if the latest round contributes no newly eligible tree. Otherwise, the current policy and parameter are retained.

\paragraph{Replay evaluation.}
Real search allows at most $BL$ search steps per round, whereas replay allows at most $2BL$ steps per historical tree. Requests that reveal a failure or have no recorded continuation both consume one replay step.

Replay quality is computed from the scores of revealed feasible nodes, including the root, using Eq.~\eqref{eq:q}. Each node contributes once, and equal scores from different nodes are counted separately. If fewer than five feasible nodes have been revealed, all available scores are averaged. Failed nodes do not contribute to the quality statistic.

\begin{figure*}[!t]
\centering
\begingroup \tcbset{colback=blue!2!white,colframe=black!55,boxrule=0.5pt, arc=2mm,left=7pt,right=7pt,top=6pt,bottom=6pt, fonttitle=\bfseries,coltitle=black,colbacktitle=blue!5!white, equal height group=discovery-prompt}
\begin{minipage}[t]{0.49\textwidth}
\begin{tcolorbox}[title=Shared Discovery Instructions]
\small\raggedright
{\color{orange!65!black}\textbf{Task}\par Design a solver for the supplied optimization problem. Propose and implement one improvement to the parent program. Preserve the required interface and task constraints.\par} \medskip {\color{blue!65!black}\textbf{Scope}\par Use only the supplied information. Do not read files, run commands, or browse. Do not choose parents, change the search policy, or modify evaluation budgets. The framework evaluates the returned program.\par} \medskip {\color{violet!75!black}\textbf{Output Format}\par Return only JSON with two strings: \texttt{hypothesis}, a concise summary of the implemented change, and \texttt{code}, the complete runnable solver with all imports and helpers. Do not return a patch or omit code.\par}
\end{tcolorbox}
\end{minipage}\hfill
\begin{minipage}[t]{0.49\textwidth}
\begin{tcolorbox}[title=Context for Each Development Request]
\small\raggedright
{\color{orange!65!black}\textbf{Task and Evaluation}\par [Problem description, required interface, and permitted changes]\par [Development instance settings, solve budgets, and resource limits]\par Maximize mean independently verified quality; every development instance must yield a feasible solution.\par} \medskip {\color{teal!65!black}\textbf{Parent and Branch History}\par Parent program: [Complete source]\par Branch history: [Earlier hypotheses, scores, and diagnostics]\par Search direction: [Assigned exploration focus]\par} \medskip {\color{blue!65!black}\textbf{Repair Context, When Needed}\par [Current source, hypothesis, error feedback, and remaining repair calls]\par Repair the same idea using the supplied feedback. Return the complete corrected program. No repair follows full development evaluation.\par}
\end{tcolorbox}
\end{minipage}
\endgroup \caption{Core discovery-agent prompt modules. Brackets denote dynamic inputs.}
\label{fig:discovery-prompt}
\end{figure*}

\paragraph{Policy proposals and selection.}
Policy learning consists of four sequential LLM calls, each requesting four distinct executable policies. Each proposed policy is replayed at $\alpha\in\{0.1,0.3,0.5,0.7,0.9\}$ in ascending order. Replay results and generation or execution failures are summarized as feedback for the next call. Valid proposals are retained even if other proposals in the same call fail; no additional repair or retry calls are made.

After all four calls, we select the valid generated policy–parameter pair with the highest replay reward. Ties favor the earlier policy proposal and, within a policy, the smaller $\alpha$. The incumbent policy and \fixed{} are also replayed to provide reference results for feedback, but are excluded from candidate selection. The incumbent is retained only if no valid generated pair is available.

\paragraph{Discovery repairs.}
The discovery agent may make up to two repair attempts for generation or compilation errors before full development evaluation. No further repairs are allowed once that evaluation begins. Every search step consumes budget, including steps that end in generation or evaluation failure. Appendix~\ref{sec:run-settings} provides the cost-accounting details.

\subsection{Core Discovery-Agent Prompt}
\label{sec:discovery-prompt}
Figure~\ref{fig:discovery-prompt} summarizes the shared instructions and request context used to develop a solver. The required interfaces, objectives, and evaluation protocols are specified in Appendix~\ref{sec:task-details}. Bracketed placeholders contain the parent program, branch-local feedback, and evaluation settings. The search policy selects the parent program, and the discovery agent proposes and implements a revision.

\subsection{Core Policy-Learning Prompt}
\label{sec:policy-prompt}
Figure~\ref{fig:policy-prompt} presents a condensed policy-learning prompt from a TSP run, separating shared instructions from context updated for each proposal call. Bracketed placeholders denote the policy source, historical search records, and replay feedback. The four calls focus in turn on stopping, branching, allocation, and revisiting.

\begin{figure*}[t]
\centering
\begingroup \tcbset{colback=blue!2!white,colframe=black!55,boxrule=0.5pt, arc=2mm,left=7pt,right=7pt,top=6pt,bottom=6pt, fonttitle=\bfseries,coltitle=black,colbacktitle=blue!5!white, equal height group=policy-prompt}
\begin{minipage}[t]{0.49\textwidth}
\begin{tcolorbox}[title=Shared Policy Instructions]
\small\raggedright
{\color{orange!65!black}\textbf{Task}\par You are the policy agent. Modify the current policy into four distinct, complete Python policies. Use observed performance to choose which branch to extend, when to open a new branch, and when to stop.\par} \medskip {\color{blue!65!black}\textbf{Interface and Observation}\par Define \texttt{policy(observation, alpha)}. Return \texttt{ROOT}, one current leaf ID, or \texttt{None}. Decisions are serial. The observation contains revealed nodes and current leaves together with the budget and action history. Per-branch statistics track gains, quality gaps, stagnation, and idle time. Use only revealed outcomes. $B$ and $L$ describe \fixed{}; the online budget is $BL$.\par} \medskip {\color{teal!65!black}\textbf{Exploration and Objective}\par The framework replays each candidate policy at 0.1, 0.3, 0.5, 0.7, and 0.9. Maximize mean $Q-\lambda A$ across historical trees, where $A$ counts replay steps. Failed and empty requests still count as replay steps.\par} \medskip {\color{violet!75!black}\textbf{Output Format}\par Return only JSON with a \texttt{policies} list of four entries. Each has \texttt{candidate\_index}, a \texttt{change} description, and complete Python \texttt{code}. No prose outside JSON.\par}
\end{tcolorbox}
\end{minipage}\hfill
\begin{minipage}[t]{0.49\textwidth}
\begin{tcolorbox}[title=Context for Each Proposal Call]
\small\raggedright
{\color{orange!65!black}\textbf{Assigned Direction}\par [Stopping / branching / allocation / revisiting]\par [Direction-specific guidance]\par For allocation: choose the route that best justifies the next search step; let promising routes continue while weak routes pause.\par} \medskip {\color{teal!65!black}\textbf{Policy and Historical Evidence}\par Current policy: [Python source and selected alpha]\par Historical evidence: [Recorded trees, per-branch search steps and gains, and replay decisions with observed outcomes]\par Replay references: [Current and incumbent results]\par} \medskip {\color{blue!65!black}\textbf{Feedback and Refinement}\par Previous proposals: [Rewards, Top5 quality, replay steps, alpha sweeps, and generation or runtime errors]\par Previous rounds: [Selected policies and replay summaries]\par Use this feedback to propose behaviorally different strategies, not alpha variants of one strategy. Keep code short; no extra repair calls.\par} \medskip {\color{violet!75!black}\textbf{Execution Context}\par [Observation schema and permitted operations]\par [$B$, $L$, online budget, replay ceiling, and fixed $\lambda$]\par No files, networks, or hidden outcomes during policy execution.\par}
\end{tcolorbox}
\end{minipage}
\endgroup \caption{Core policy-learning prompt modules. Brackets denote dynamic inputs.}
\label{fig:policy-prompt}
\end{figure*}

Each subsequent call receives replay results and generation or execution failures from earlier calls. The policy supplied for refinement is the valid proposal with the highest replay reward so far in the current update. If no valid proposal is available, the incumbent is supplied instead. The policy agent uses completed historical trees to guide proposal generation. During execution, each proposed policy receives observations containing only outcomes revealed by its own actions.

\subsection{An Example of a Learned Policy}
\label{sec:policy-example}
Listing~\ref{lst:policy-example} shows a learned policy from the GPT-6-Luna TSP experiment. It was selected after the eighth search round and deployed in the ninth with $\alpha=0.9$. The ninth round produced the program with the highest development score in the run. The code is reproduced with only line wrapping changed. Appendix~\ref{sec:policy-trajectories} follows its decisions and the resulting real search trajectory.

The policy balances continuing recent progress with revisiting previously promising branches. It continues branches that show recent gains, little stagnation, and a small gap to the best observed quality. It also revisits paused branches whose earlier gains and current quality justify further exploration. Eligible branches are ranked by a score that rewards gains and penalizes stagnation and quality gaps. If the highest-ranked branch exceeds the continuation threshold, the policy extends it. Otherwise, it opens a new branch if fewer than $B$ branches exist, or stops the round. This learned policy therefore retains $B$ as its branching cap while adapting branch lengths and revisits to observed progress.

Increasing $\alpha$ lowers the recent-gain and continuation thresholds and relaxes quality-gap limits. For revisiting a branch, it also permits more stagnation while requiring a longer interval since the branch's last request.

Here, \texttt{recent\_gains} contains up to three most recent increases in a branch's best score, and \texttt{gap\_to\_best} measures how far that score trails the global best. \texttt{stagnation} counts requests to the branch since its last improvement. \texttt{idle\_batches} counts completed serial search steps since the branch's last request. All quantities are computed from revealed outcomes.

\raggedbottom

\section{Benchmark Tasks, Interfaces, and Metrics}
\label{sec:task-details}
\begin{figure*}[!t]
\centering
\begin{minipage}{0.96\textwidth}
\begin{lstlisting}[
  style=dreamprogram,
  caption={A learned TSP search policy, deployed with $\alpha=0.9$.},
  label={lst:policy-example}
]
def policy(observation, alpha):
    if (observation["remaining"] <= 0
            or observation["used"] >= observation["budget"]):
        return None
    leaves = observation.get("leaves", [])
    options = []
    for r in observation.get("branch_stats", []):
        leaf = next((n for n in leaves
                     if n["id"] == r["leaf_id"]), None)
        if leaf is None or leaf.get("status") == "interrupted":
            continue
        gains = r.get("recent_gains", [])
        peak = max(gains) if gains else 0.0
        recent = sum(gains) / max(1, len(gains))
        stale = r.get("stagnation", 0)
        gap = r.get("gap_to_best", 0.0)
        idle = r.get("idle_batches", 0)
        delayed = (peak >= 0.001 and idle >= 1 + int(3 * alpha)
                   and stale <= 2 + int(2 * alpha)
                   and gap <= 0.004 + 0.008 * alpha)
        active = (recent >= 0.0003 * (1.0 - alpha) + 0.00004
                  and stale <= 1 and gap <= 0.003 + 0.007 * alpha)
        if delayed or active:
            value = max(peak / (2.0 + stale) if delayed else 0.0,
                        recent / (1.0 + stale)) - 0.04 * gap
            options.append((value, idle if delayed else -idle, leaf["id"]))
    if options:
        value, _, leaf_id = max(options)
        if value > 0.00012 * (1.0 - alpha):
            return leaf_id
    if observation["branches"] < observation["B"]:
        return "ROOT"
    return None
\end{lstlisting}
\end{minipage}
\end{figure*}

\paragraph{Reference gaps.}
For objective $z_i$ and reference $b_i$ on instance $i$, the mean relative gap over $M$ instances is
\begin{equation}
 \operatorname{Gap}=\frac{100}{M}\sum_{i=1}^{M}
 \begin{cases}
  (z_i-b_i)/b_i,&\text{minimization},\\
  (b_i-z_i)/b_i,&\text{maximization}.
 \end{cases}
 \label{eq:gap}
\end{equation}
We compute gaps relative to LKH tour lengths for TSP, certified OR-Tools optima for KP, the volume lower bound for OBPP, and the supplied Taillard upper bounds for FSSP. A negative gap of FSSP indicates a makespan below the reference upper bound ($C_{\max}<b_i$).

\paragraph{Traveling salesman problem (TSP).}
The evolved selector implements \texttt{select\_next\_node(} \texttt{current\_node, destination\_node, unvisited\_nodes, distance\_matrix)}. It receives the current city, the return destination (city zero), the unvisited cities, and the matrix of unrounded Euclidean distances, and returns one unvisited city ID. The fixed constructor starts at city zero and repeatedly invokes the selector until all cities have been visited, then closes the tour.

The development set consists of 64 50-city instances in \texttt{train50\_dataset.npy} from the public MCTS-AHD dataset.\footnote{ \url {https://github.com/zz1358m/MCTS-AHD-master/tree/main/problems/tsp_constructive/dataset}} For each 50-city instance, a tour of length $\ell$ gives quality $1-\ell/(50\sqrt{2})$. All reported methods and reference solvers are evaluated on the same 4,000 synthetic instances at $N\in\{50,100,200,500\}$ and 15 TSPLIB instances, totaling 4,015 test instances. Gaps are computed relative to LKH tour lengths, which match the official optima on all 15 TSPLIB instances (Appendix~\ref{sec:tsp-extension}).

\paragraph{Knapsack problem (KP).}
The evolved selector implements \texttt{select\_next\_item(} \texttt{remaining\_capacity, weights, values)}. It receives the remaining knapsack capacity and the weight and value arrays of the remaining items, and returns the index of an item that fits within the remaining capacity. The index refers to the supplied arrays. The fixed constructor removes the selected item, updates the remaining capacity, and repeats until no remaining item fits. The selected subset is checked against the original instance.

The development set consists of 64 instances at $N=100$ in \texttt{train100\_dataset.npy} from the public MCTS-AHD dataset.\footnote{ \url {https://github.com/zz1358m/MCTS-AHD-master/tree/main/problems/kp_constructive/dataset}} Item weights and values are uniformly distributed in $[0,1]$. The knapsack capacity is $W=12.5$ for $N=50$ items and $W=25$ for $N\in\{100,200,500\}$. The development score is the total value of the selected items divided by 100. Test results report the total selected value without normalization. The official test checker uses a feasibility tolerance of $10^{-6}$. The OR-Tools reference enforces the original capacities without this tolerance.

\paragraph{Online bin packing problem (OBPP).}
The evolved function implements \texttt{score(item, bins)}, where \texttt{item} is the arriving item's size and \texttt{bins} contains the remaining capacities of all feasible bins, including unused bins. It returns one score for each supplied bin. The fixed controller places the item in the bin with the highest score, breaking ties by choosing the first bin in the supplied order.

We use the released MCTS-AHD dataset for OBPP.\footnote{ \url{https://github.com/zz1358m/MCTS-AHD-master/tree/main/problems/bpp_online/dataset}} The four development instances cover the combinations of $N\in\{1,000,5,000\}$ items and bin capacities $C\in\{100,500\}$. The 30 test instances additionally include $N=10,000$ at both capacities. For an instance with item weights $w_i$, and $K$ used bins, the volume lower bound is $b=\lceil\sum_i w_i/C\rceil$. The development score is $b/K$, averaged equally across the four development instances. Test performance is the percentage excess in bin count over this bound, $100(K/b-1)$, averaged equally across the 30 test instances.

\paragraph{Flow-shop scheduling problem (FSSP).}
The evolved perturbation function implements \texttt{get\_matrix\_and\_jobs(current\_sequence, time\_matrix, m, n)}. Here, \texttt{current\_sequence} lists the current job order, \texttt{time\_matrix} is the $n\times m$ processing-time matrix, and \texttt{n} and \texttt{m} are the numbers of jobs and machines, respectively. The function returns a modified processing-time matrix of the same shape and a list of job IDs to perturb. The fixed controller initializes the sequence with NEH and uses these outputs to guide swap and insertion perturbations within GLS. Makespan is computed from the original processing times.

The development set consists of all 64 instances in the public EoH training set, with 50 jobs and 2--20 machines. \footnote{ \url{https://github.com/FeiLiu36/EoH/tree/main/examples/fssp_gls_numba/TrainingData}} The development score is $1-C_{\max}/7000$. Testing uses the 110 Taillard instances, with ten instances per job--machine size. The supplied upper bounds serve as reference makespans.

GLS is allowed up to 30 seconds during development and 60 seconds during testing, with a maximum of 1,000 iterations in both phases. Reaching the iteration cap can therefore end a run before its time budget is exhausted. 

\paragraph{Two-dimensional bin packing problem (2D-BPP).}
The evolved program implements \texttt{solve(instance, time\_budget, seed)}. The \texttt{instance} provides the bin dimensions and rectangle sizes, \texttt{time\_budget} specifies the solve time allowance in seconds, and \texttt{seed} controls randomness. The program returns one $(\mathrm{bin},x,y)$ placement per rectangle in input order, specifying its bin ID and lower-left coordinates. The checker verifies integer coordinates, preservation of the original orientations, bin containment, and pairwise non-overlap.

The development set consists of 60 instances generated following the BWMV class distributions, with a 20-second solve budget per instance. The development score is $K^{-1}\sum_{b=1}^{K}u_b^2$, where $K$ is the number of used bins and $u_b$ is the area utilization of bin b. Testing uses 500 BWMV instances with a 60-second budget per instance and reports the number of used bins. \footnote{ \url{https://github.com/pesho-ivanov/glass-cutting/tree/master/data/BWMV(2dbpp)}} These instances cover ten problem classes and five sizes from 20 to 100 items in increments of 20, with ten instances per class--size combination.

The reference results use the aNB variant of BRKGA without rotation \citep{goncalves2013brkga}, with equal weight assigned to each class-–size combination. Table~\ref{tab:fullsolver} reports the BRKGA bin counts and the excess bin counts of \fixed{} and \dream{} relative to this reference. Each size row sums over 100 instances, and the All row sums over all 500 instances.

\paragraph{Local reference solvers.}
Greedy uses nearest-neighbor construction from city zero for TSP and selects feasible items by value-to-weight ratio for KP. POMO is evaluated locally using the official TSP50 (Python version) and KP200 (legacy version) checkpoints fixed before testing. We use greedy decoding from a single start at index zero without augmentation, following the MCTS-AHD setting \citep{zheng2025mcts}. POMO runs on NVIDIA RTX PRO 6000 Blackwell GPUs. CPU reference evaluations run on a dual-socket AMD EPYC 9555 machine.

LKH-3.0.13 uses ten runs, with at most 1,000 trials and a 60-second limit per run. \footnote{ \url{http://webhotel4.ruc.dk/~keld/research/LKH-3/}} Distances are scaled by $10^6$ and rounded for search, while reported tour lengths are recomputed from the original coordinates.

OR-Tools 9.14 uses its multidimensional branch-and-bound knapsack solver with one capacity constraint and a 60-second limit. \footnote{ \url{https://developers.google.com/optimization/pack/knapsack}} The stored binary inputs are scaled by $2^{53}$ to obtain exact integer representations. Assertions verify exact representability and compliance with the signed 64-bit range. OR-Tools certifies optimality for every evaluated KP instance.

\section{Experimental Details and Cost Accounting}
\label{sec:run-settings}
For each task and search method, we select the solver with the highest development score in each run and evaluate it directly on the test set.

\paragraph{Baseline coverage.}
Baseline coverage is determined by the problem-specific implementations publicly released by the original authors. Accordingly, we include EoH, ReEvo, and MCTS-AHD for constructive TSP; MCTS-AHD for KP; EoH for FSSP; and FunSearch, EoH, ReEvo, and MCTS-AHD for OBPP. Model names indicate the LLM backbones used for algorithm discovery.

\paragraph{Rounds and execution.}
For root generation in each round, the discovery agent receives only the same task specification and reference program. Evaluation caches are maintained separately for each round. Search experience therefore transfers across rounds through the learned policy. In the paired GPT-6-Luna TSP and 2D-BPP comparisons, \fixed{} and \dream{} share the initial solver. \fixed{} permits up to $W$ concurrent development requests, whereas \dream{} issues one request at a time.

\paragraph{Cost accounting.}
Token counts cover recorded LLM usage for solver development and policy learning, including root generation and repair calls. Invalid and failed search steps still consume the search budget. A single search step may involve multiple LLM calls when repairs are needed, so we report search steps and token usage separately.

\paragraph{Result aggregation.}
All reported metrics are computed from complete per-instance evaluation results. Reference gaps are calculated for each instance before averaging. For results reported over three runs, we average the run-level means with equal weight. In particular, the overall GPT-6-Luna TSP gap averages all 4,015 test instances equally within each run, then averages across the three runs. Table highlighting is based on values rounded to the displayed precision.

\paragraph{Development-curve aggregation.}
Figures~\ref{fig:tsp-evolution} and~\ref{fig:2dbpp-generation} show the mean best-so-far development score across three runs at each search step. Invalid candidates and failed evaluations leave each run's best score unchanged. Within each method, the final best scores of runs that end earlier are carried forward until the longest run ends. Shading indicates $\pm1$ standard error across the three runs.

\FloatBarrier

\section{Extended Results}
\label{sec:extended-results}
This appendix supplements Section~\ref{sec:experiments} with detailed comparisons and analyses of search behavior. We first report TSP solution quality and running times, FSSP results alongside CP baselines, and OBPP performance across item counts and bin capacities. We then examine the effects of policy replay and the cost penalty on 2D-BPP, including paired comparisons and token expenditure. Finally, we use real search trajectories to illustrate how learned policies extend promising branches, revisit alternatives, and decide when to stop.

\subsection{TSP Solution Quality and Running Times}
\label{sec:tsp-times}
Table~\ref{tab:tsp-times} reports mean reference gaps and per-instance solve times for the same solvers and test instances as Table~\ref{tab:constructive}. GPT-6-Luna results average three runs per method. With this backbone, \dream{} achieves lower mean gaps and shorter mean solve times than \fixed{} in all five test groups. Reported times measure solver evaluation. For LLM-generated solvers, timing covers only the solver call and excludes process startup and CPU queueing.

\begin{table*}[!t]
\centering\small
\setlength{\tabcolsep}{3pt}
\resizebox{1.0\textwidth}{!}{%
\begin{tabular}{@{}llrrrrrrrrrr@{}}
\toprule
\multirow{2}{*}{LLM Model} & \multirow{2}{*}{Method} &
  \multicolumn{2}{c}{$N=50$} & \multicolumn{2}{c}{$N=100$} &
  \multicolumn{2}{c}{$N=200$} & \multicolumn{2}{c}{$N=500$} &
  \multicolumn{2}{c}{TSPLIB} \\
 & & Gap (\%) & Time (s) & Gap (\%) & Time (s) & Gap (\%) & Time (s) &
  Gap (\%) & Time (s) & Gap (\%) & Time (s) \\
\midrule
 & LKH ref. & -- & 0.479 & -- & 1.83 & -- & 6.06 & -- & 31.3 & -- & 29.1 \\
 & Greedy & 22.622 & 0.0002 & 24.936 & 0.0005 & 25.711 & 0.0013 & 25.903 & 0.0239 &
  24.332 & 0.0065 \\
 & POMO$^{\dagger}$ & \textbf{0.383} & 0.0005 & \textbf{2.999} & 0.0004 & 20.447 &
  0.0008 & 46.135 & 0.0024 & 23.923 & 0.0502 \\
\midrule
\multirow{5}{*}{GPT-4o-mini} & EoH & 16.822 & 0.001 & 19.114 & 0.0025 & 20.842 &
  0.0085 & 22.209 & 0.0502 & 21.060 & 0.0096 \\
 & ReEvo & 22.622 & 0.0006 & 24.936 & 0.0014 & 25.711 & 0.005 & 25.903 & 0.0467 &
  24.332 & 0.0084 \\
 & MCTS-AHD & 15.906 & 0.0049 & 18.832 & 0.0203 & 21.166 & 0.0858 & 22.549 & 0.527 &
  19.306 & 0.0936 \\
 & \fixed{} & 22.622 & 0.0006 & 24.936 & 0.0013 & 25.711 & 0.0044 & 25.903 & 0.0528 &
  24.332 & 0.0095 \\
 & \method{} & \cellcolor{gray!18}14.379 & 0.0006 & \cellcolor{gray!18}15.952 &
  0.0013 & \cellcolor{gray!18}17.557 & 0.0044 & \cellcolor{gray!18}19.149 & 0.0526 &
  \cellcolor{gray!18}17.390 & 0.0105 \\
\midrule
\multirow{5}{*}{GPT-6-Luna} & EoH & 16.819 & 0.0716 & 18.955 & 0.599 & 20.305 & 5.32 &
  21.099 & 98.3 & 20.658 & 12.2 \\
 & ReEvo & 15.682 & 0.602 & 18.493 & 0.62 & 20.292 & 0.834 & 21.440 & 5.04 & 20.782 &
  1.01 \\
 & MCTS-AHD & 15.959 & 0.21 & 17.929 & 1.92 & 19.143 & 18.5 & 19.728 & 488 & 20.226 &
  64 \\
 & \fixed{} & 15.275 & 0.0807 & 17.432 & 0.668 & 18.495 & 6.21 & 19.378 & 137 &
  18.355 & 14.5 \\
 & \method{} & \cellcolor{gray!18}9.489 & 0.0622 & \cellcolor{gray!18}11.316 & 0.398 &
  \cellcolor{gray!18}\textbf{12.361} & 3.56 & \cellcolor{gray!18}\textbf{13.209} &
  80.3 & \cellcolor{gray!18}\textbf{11.984} & 10.3 \\
\bottomrule
\end{tabular}
} \caption{TSP mean gaps (\%) to LKH and per-instance solve times (seconds). Bold indicates the lowest gap among non-reference methods. Shading indicates the lowest gap within each LLM backbone, including ties at displayed precision. $^{\dagger}$POMO's synthetic-instance times are GPU batch inference times divided by batch size.
}
\label{tab:tsp-times}
\end{table*}

POMO runs on the GPU, while the other methods run on CPUs. For synthetic instances, POMO's per-instance time is the synchronized batch inference time divided by the actual batch size, with up to 60 instances per batch. TSPLIB instances are evaluated individually because their sizes vary. Reported times correspond to the tours used for gap calculation and are averaged equally over instances within each test group. GPT-6-Luna results then average the three run-level means. Times are reported with up to three significant figures and at most four decimal places.

\paragraph{Larger and structured instances.}
\label{sec:tsp-extension}
The selected solvers are evaluated on all 1,000 released 500-city MCTS-AHD instances and the 15 TSPLIB instances listed in Table~\ref{tab:tsplib-results}. Each constructor runs to completion without a CPU or wall-clock cutoff. Evaluation permits at most 50 concurrent instances per solver.

Before inference, TSPLIB coordinates are translated by subtracting the minimum on each axis and uniformly scaled by the larger of the two coordinate ranges. Inference uses a single start at city zero without augmentation. Returned tours are evaluated on the original coordinates using TSPLIB EUC\_2D nearest-integer edge costs. LKH uses the same edge costs. Gaps are computed relative to LKH tour lengths, which match the official optima on all 15 instances. \footnote{\url{https://comopt.ifi.uni-heidelberg.de/software/TSPLIB95/tsp/TSP-BEST.html}}

\begin{table*}[!t]
\centering\small
\setlength{\tabcolsep}{2pt}
\resizebox{\textwidth}{!}{%
\begin{tabular}{@{}lrr*{7}{rr}@{}}
\toprule
\multirow{2}{*}{Instance} & \multirow{2}{*}{$N$} & \multirow{2}{*}{LKH (ref.)} &
  \multicolumn{2}{c}{Greedy} & \multicolumn{2}{c}{POMO} & \multicolumn{2}{c}{EoH} &
  \multicolumn{2}{c}{ReEvo} & \multicolumn{2}{c}{MCTS-AHD} &
  \multicolumn{2}{c}{\fixed{}} & \multicolumn{2}{c}{\method{}} \\
 &  &  & Length & Gap (\%) & Length & Gap (\%) & Length & Gap (\%) & Length &
  Gap (\%) & Length & Gap (\%) & Length & Gap (\%) & Length & Gap (\%) \\
\midrule
ts225 & 225 & 126643 & 162697.0 & 28.469 & 182934.0 & 44.449 & 145719.7 & 15.063 &
  146561.7 & 15.728 & 152030.7 & 20.047 & 148299.7 & 17.101 &
  \cellcolor{gray!18}\textbf{141445.0} & \cellcolor{gray!18}\textbf{11.688} \\
rat99 & 99 & 1211 & 1558.0 & 28.654 & \textbf{1332.0} & \textbf{9.992} & 1424.7 &
  17.644 & 1437.7 & 18.717 & 1544.0 & 27.498 & 1456.0 & 20.231 &
  \cellcolor{gray!18}1355.7 & \cellcolor{gray!18}11.946 \\
bier127 & 127 & 118282 & 135848.0 & 14.851 & 134250.0 & 13.500 & 150183.0 & 26.970 &
  133371.3 & 12.757 & 137271.7 & 16.055 & 132660.3 & 12.156 &
  \cellcolor{gray!18}\textbf{127479.7} & \cellcolor{gray!18}\textbf{7.776} \\
lin318 & 318 & 42029 & 54019.0 & 28.528 & 59945.0 & 42.628 & 50308.0 & 19.698 &
  53587.7 & 27.502 & 51346.7 & 22.170 & 51914.0 & 23.519 &
  \cellcolor{gray!18}\textbf{48487.0} & \cellcolor{gray!18}\textbf{15.366} \\
eil51 & 51 & 426 & 511.0 & 19.953 & \textbf{429.0} & \textbf{0.704} & 487.0 & 14.319 &
  \cellcolor{gray!18}475.3 & \cellcolor{gray!18}11.581 & 487.0 & 14.319 & 513.3 &
  20.501 & 479.7 & 12.598 \\
d493 & 493 & 35002 & 43395.0 & 23.979 & 55695.0 & 59.119 & 41895.3 & 19.694 &
  40781.0 & 16.510 & 41178.0 & 17.645 & 41451.7 & 18.427 &
  \cellcolor{gray!18}\textbf{38873.0} & \cellcolor{gray!18}\textbf{11.059} \\
kroB100 & 100 & 22141 & 29158.0 & 31.692 & \textbf{22904.0} & \textbf{3.446} &
  26536.3 & 19.852 & 27662.7 & 24.939 & 26526.0 & 19.805 & 26782.3 & 20.963 &
  \cellcolor{gray!18}24791.0 & \cellcolor{gray!18}11.969 \\
kroC100 & 100 & 20749 & 26327.0 & 26.883 & 22999.0 & 10.844 & 24969.3 & 20.340 &
  24882.7 & 19.922 & 25009.3 & 20.533 & 23857.3 & 14.981 &
  \cellcolor{gray!18}\textbf{22639.3} & \cellcolor{gray!18}\textbf{9.110} \\
ch130 & 130 & 6110 & 7578.0 & 24.026 & \textbf{6530.0} & \textbf{6.874} & 7073.7 &
  15.772 & 7319.0 & 19.787 & 7052.0 & 15.417 & 6966.0 & 14.010 &
  \cellcolor{gray!18}6914.0 & \cellcolor{gray!18}13.159 \\
pr299 & 299 & 48191 & 62738.0 & 30.186 & 70010.0 & 45.276 & 59506.3 & 23.480 &
  58868.0 & 22.156 & 58805.3 & 22.026 & 57259.3 & 18.817 &
  \cellcolor{gray!18}\textbf{53958.3} & \cellcolor{gray!18}\textbf{11.968} \\
fl417 & 417 & 11861 & 14598.0 & 23.076 & 15072.0 & 27.072 & 15062.3 & 26.990 &
  16228.7 & 36.824 & 14634.3 & 23.382 & 15013.0 & 26.574 &
  \cellcolor{gray!18}\textbf{13552.3} & \cellcolor{gray!18}\textbf{14.260} \\
kroA150 & 150 & 26524 & 33612.0 & 26.723 & \textbf{29279.0} & \textbf{10.387} &
  31133.3 & 17.378 & 32189.7 & 21.361 & 31163.7 & 17.492 & 30266.7 & 14.110 &
  \cellcolor{gray!18}29894.0 & \cellcolor{gray!18}12.705 \\
pr264 & 264 & 49135 & 57963.0 & 17.967 & 63174.0 & 28.572 & 60580.0 & 23.293 &
  58899.7 & 19.873 & 58717.0 & 19.501 & 58391.7 & 18.839 &
  \cellcolor{gray!18}\textbf{55638.7} & \cellcolor{gray!18}\textbf{13.236} \\
pr226 & 226 & 80369 & 94277.0 & 17.305 & \textbf{87381.0} & \textbf{8.725} & 99982.0 &
  24.404 & 96235.7 & 19.742 & 97972.3 & 21.903 & 89031.7 & 10.779 &
  \cellcolor{gray!18}88338.3 & \cellcolor{gray!18}9.916 \\
pr439 & 439 & 107217 & 131549.0 & 22.694 & 157886.0 & 47.258 & 133991.7 & 24.972 &
  133310.7 & 24.337 & 134668.0 & 25.603 & 133283.0 & 24.311 &
  \cellcolor{gray!18}\textbf{121153.0} & \cellcolor{gray!18}\textbf{12.998} \\
\midrule
Mean & -- & 46392.7 & 57055.2 & 24.332 & 60654.7 & 23.923 & 56590.2 & 20.658 &
  55454.1 & 20.782 & 55893.7 & 20.226 & 54476.4 & 18.355 &
  \cellcolor{gray!18}\textbf{51666.6} & \cellcolor{gray!18}\textbf{11.984} \\
\bottomrule
\end{tabular}
} \caption{TSPLIB tour lengths and gaps (\%) to LKH, which matches the optima on all 15 instances. LLM-based results use GPT-6-Luna and average three runs per method. Bold indicates the best non-reference result. Shading indicates the best LLM-based result, including ties at displayed precision.
}
\label{tab:tsplib-results}
\end{table*}

\FloatBarrier
\Needspace{6\baselineskip}
\subsection{FSSP Evaluation and CP Solver Settings}
\label{sec:fssp-makespan}
Table~\ref{tab:fssp-makespan} reports mean makespan and per-instance solve time for the FSSP methods, including the two CP baselines. \dream{} achieves the lowest mean makespan among the LLM-based methods with both GPT-4o-mini and GPT-6-Luna.

\begin{table*}[!t]
\centering\small
\setlength{\tabcolsep}{2pt}
\resizebox{\textwidth}{!}{%
\begin{tabular}{@{}rrr*{8}{rr}@{}}
\toprule
\multicolumn{7}{c}{} & \multicolumn{6}{c}{GPT-4o-mini}
& \multicolumn{6}{c}{GPT-6-Luna} \\
\cmidrule(lr){8-13}\cmidrule(lr){14-19}
$N$ & $M$ & Taillard UB & \multicolumn{2}{c}{CP-SAT}
& \multicolumn{2}{c}{OptalCP} & \multicolumn{2}{c}{EoH}
& \multicolumn{2}{c}{\fixed{}} & \multicolumn{2}{c}{\method{}}
& \multicolumn{2}{c}{EoH} & \multicolumn{2}{c}{\fixed{}}
& \multicolumn{2}{c}{\method{}} \\
\cmidrule(lr){4-5}\cmidrule(lr){6-7}\cmidrule(lr){8-9}
\cmidrule(lr){10-11}\cmidrule(lr){12-13}\cmidrule(lr){14-15}
\cmidrule(lr){16-17}\cmidrule(lr){18-19}
& & $C_{\max}$ & $C_{\max}$ & Time & $C_{\max}$ & Time
& $C_{\max}$ & Time & $C_{\max}$ & Time
& $C_{\max}$ & Time & $C_{\max}$ & Time
& $C_{\max}$ & Time & $C_{\max}$ & Time \\
\midrule
20 & 5 & 1222.5 & \textbf{1221.9} & 5.87 & \textbf{1221.9} & 270 & 1223.6 & 0.280
& 1223.6 & 0.282 & \cellcolor{gray!18}1222.4 & 0.270 & \cellcolor{gray!18}1222.4
& 0.356 & \cellcolor{gray!18}1222.4 & 2.09 & \cellcolor{gray!18}1222.4 & 0.536 \\
20 & 10 & 1513.7 & 1516.5 & 245 & 1521.1 & 300 & 1516.2 & 0.330 & 1516.2 & 0.327
& \cellcolor{gray!18}1515.5 & 0.313 & 1515.2 & 0.444 & 1521.1 & 2.68
& \cellcolor{gray!18}\textbf{1514.2} & 0.639 \\
20 & 20 & 2235.1 & 2246.8 & 300 & 2253.3 & 300 & \cellcolor{gray!18}2237.9 & 0.494
& \cellcolor{gray!18}2237.9 & 0.501 & 2239.6 & 0.438 & 2237.1 & 0.691 & 2239.3
& 4.34 & \cellcolor{gray!18}\textbf{2236.1} & 0.920 \\
50 & 5 & 2736.4 & \textbf{2736.4} & 13.7 & \textbf{2736.4} & 182 & 2736.5 & 2.19
& 2736.5 & 2.23 & \cellcolor{gray!18}\textbf{2736.4} & 1.94
& \cellcolor{gray!18}\textbf{2736.4} & 2.21 & 2736.6 & 5.12 & 2736.7 & 2.51 \\
50 & 10 & 2997.6 & 3037.8 & 300 & 3011.7 & 300 & \cellcolor{gray!18}3000.1 & 2.77
& \cellcolor{gray!18}3000.1 & 2.95 & 3001.5 & 2.55 & 2998.6 & 2.99 & 3000.7 & 3.87
& \cellcolor{gray!18}\textbf{2997.5} & 3.16 \\
50 & 20 & 3731.0 & 3943.0 & 300 & 3854.4 & 300 & \cellcolor{gray!18}3756.3 & 4.78
& \cellcolor{gray!18}3756.3 & 5.19 & 3761.6 & 4.67 & 3751.8 & 5.64 & 3761.1 & 6.90
& \cellcolor{gray!18}\textbf{3750.4} & 5.47 \\
100 & 5 & 5246.9 & 5246.4 & 153 & 5245.5 & 300 & 5245.0 & 12.5 & 5245.0 & 13.7
& \cellcolor{gray!18}\textbf{5244.5} & 13.5 & 5244.8 & 14.8 & 5244.8 & 26.1
& \cellcolor{gray!18}\textbf{5244.5} & 13.9 \\
100 & 10 & 5630.0 & 5752.0 & 300 & 5652.6 & 300 & 5637.9 & 18.5 & 5637.9 & 18.8
& \cellcolor{gray!18}5635.2 & 19.3 & 5641.0 & 20.4
& \cellcolor{gray!18}\textbf{5631.2} & 33.7 & 5631.3 & 19.9 \\
100 & 20 & 6362.6 & 6634.4 & 300 & 6538.7 & 300 & \cellcolor{gray!18}6380.3 & 37.3
& \cellcolor{gray!18}6380.3 & 35.1 & 6384.4 & 36.2 & 6369.0 & 34.0 & 6375.3 & 40.8
& \cellcolor{gray!18}\textbf{6350.8} & 39.0 \\
200 & 10 & 10673.0 & 10803.8 & 300 & 10755.5 & 300 & 10700.4 & 60 & 10700.4 & 60
& \cellcolor{gray!18}\textbf{10676.6} & 60 & 10687.5 & 60 & 10682.4 & 60
& \cellcolor{gray!18}10677.3 & 60 \\
200 & 20 & 11373.6 & 11752.4 & 300 & 11726.5 & 300 & 11440.3 & 60 & 11430.9 & 60
& \cellcolor{gray!18}11411.7 & 60 & 11441.2 & 60 & 11406.2 & 60
& \cellcolor{gray!18}\textbf{11376.2} & 60 \\
\midrule
\multicolumn{2}{l}{Mean} & 4883.9 & 4990.1 & 229 & 4956.1 & 287 & 4897.7 & 18.1
& 4896.8 & 18.1 & \cellcolor{gray!18}4893.6 & 18.1 & 4895.0 & 18.3 & 4892.8 & 22.4
& \cellcolor{gray!18}\textbf{4885.2} & 18.8 \\
\bottomrule
\end{tabular}
} \caption{FSSP mean makespan and per-instance solve time (seconds) for $N$ jobs and $M$ machines. Bold indicates the lowest makespan among non-reference methods. Shading indicates the lowest makespan within each LLM backbone, including ties at displayed precision.
}
\label{tab:fssp-makespan}
\end{table*}

We choose OR-Tools CP-SAT 9.14.6206 and OptalCP 2026.4.0 as CP baselines. Both enforce the permutation flow-shop constraints: every job visits the machines in order, operations on each machine cannot overlap, and all machines follow the same job permutation. CP-SAT represents the shared job order using pairwise ordering variables across machines. OptalCP enforces it through a same-sequence constraint on the machine sequence variables. Both minimize makespan and use the same deterministic NEH solution as a warm start. The supplied Taillard reference upper bounds are used only for reporting.

Each CP solver uses one thread per instance and has a 300-second time limit. Timing starts after model construction and NEH initialization. Every returned permutation is independently checked, and its makespan is recomputed using the original processing times. Both solvers return feasible solutions on all 110 instances. CP-SAT proves optimality on 30 instances, and OptalCP on five.

\paragraph{Paired \fixed{}/\dream{} comparison.}
\label{sec:fssp-paired}
For each LLM backbone, we pair \fixed{} and \dream{} makespans by instance ID on all 110 instances under the same 60-second GLS limit. A win for \dream{} means a lower makespan than \fixed{} on the same instance; ties require exact equality of integer makespans. With GPT-6-Luna, \dream{} records 59 wins, 40 ties, and 11 losses, with a mean makespan reduction of 7.61. With GPT-4o-mini, \dream{} yields 38 wins, 46 ties, and 26 losses, with a mean reduction of 3.25. Excluding ties, separate two-sided exact sign tests of equal win and loss probabilities give $p=4.47\times10^{-9}$ on 70 non-tied instances for GPT-6-Luna and $p=0.169$ on 64 for GPT-4o-mini. 

\FloatBarrier
\Needspace{6\baselineskip}
\subsection{Online Bin Packing Results}
\label{sec:obpp-results}
Table~\ref{tab:obpp-details} breaks down OBPP performance by item count and bin capacity. \dream{} achieves its largest gains over \fixed{} on longer sequences at $C=100$. At $N=10,000$, it reduces the mean gap from 3.501\% to 0.459\%, using 122 fewer bins per instance on average. At $C=500$, the methods obtain similar bin counts close to the volume lower bound.

\begin{table*}[!htbp]
\centering\small
\setlength{\tabcolsep}{3pt}
\resizebox{\textwidth}{!}{%
\begin{tabular}{@{}rr*{7}{rr}@{}}
\toprule
\multirow{2}{*}{$N$} & \multirow{2}{*}{$C$} & \multicolumn{2}{c}{Ref. LB}
& \multicolumn{2}{c}{FunSearch} & \multicolumn{2}{c}{EoH}
& \multicolumn{2}{c}{ReEvo} & \multicolumn{2}{c}{MCTS-AHD}
& \multicolumn{2}{c}{\fixed{}} & \multicolumn{2}{c}{\method{}} \\
\cmidrule(lr){3-4}\cmidrule(lr){5-6}\cmidrule(lr){7-8}\cmidrule(lr){9-10}
\cmidrule(lr){11-12}\cmidrule(lr){13-14}\cmidrule(lr){15-16}
& & Bins & Gap (\%) & Bins & Gap (\%) & Bins & Gap (\%) & Bins & Gap (\%)
& Bins & Gap (\%) & Bins & Gap (\%) & Bins & Gap (\%) \\
\midrule
1k & 100 & 402.4 & - & 421.6 & 4.771 & 421.6 & 4.771 & 420.6 & 4.522
& \cellcolor{gray!18}418.2
& \cellcolor{gray!18}3.926 & 420.0 & 4.371 & 420.4 & 4.477 \\
1k & 500 & 80.6 & - & \cellcolor{gray!18}80.8 & \cellcolor{gray!18}0.247
& \cellcolor{gray!18}80.8 & \cellcolor{gray!18}0.247 & \cellcolor{gray!18}80.8
& \cellcolor{gray!18}0.247 & \cellcolor{gray!18}80.8 & \cellcolor{gray!18}0.247
& \cellcolor{gray!18}80.8 & \cellcolor{gray!18}0.247 & \cellcolor{gray!18}80.8
& \cellcolor{gray!18}0.247 \\
\addlinespace
5k & 100 & 2019.4 & - & 2106.4 & 4.307 & 2106.4 & 4.307 & 2095.0 & 3.743 & 2087.2 & 3.357
& 2091.8
& 3.585 & \cellcolor{gray!18}2041.4 & \cellcolor{gray!18}1.089 \\
5k & 500 & 402.4 & - & 404.6 & 0.547 & 404.6 & 0.547 & 404.6 & 0.547 & 404.6 & 0.547
& 404.4
& 0.497 & \cellcolor{gray!18}404.2 & \cellcolor{gray!18}0.448 \\
\addlinespace
10k & 100 & 4010.6 & - & 4173.0 & 4.049 & 4173.0 & 4.049 & 4160.2 & 3.730 & 4142.6 & 3.291
& 4151.0 & 3.501 & \cellcolor{gray!18}4029.0 & \cellcolor{gray!18}0.459 \\
10k & 500 & 802.4 & - & \cellcolor{gray!18}806.2 & \cellcolor{gray!18}0.474
& \cellcolor{gray!18}806.2 & \cellcolor{gray!18}0.474 & \cellcolor{gray!18}806.2
& \cellcolor{gray!18}0.474 & \cellcolor{gray!18}806.2 & \cellcolor{gray!18}0.474
& \cellcolor{gray!18}806.2 & \cellcolor{gray!18}0.474 & \cellcolor{gray!18}806.2
& \cellcolor{gray!18}0.474 \\
\bottomrule
\end{tabular}
} \caption{OBPP mean bin counts and gaps (\%) to the volume lower bound by item count $N$ and bin capacity $C$. All LLM-based methods use GPT-6-Luna. Each row averages five instances. Ref. LB reports the mean volume lower bound. Shading marks the lowest non-reference values in each row and metric, including ties at displayed precision.
}
\label{tab:obpp-details}
\end{table*}

\FloatBarrier
\Needspace{6\baselineskip}
\subsection{Comparing Replay Quality Measures}
\label{sec:replay-quality-ablation}

To complement the analysis of delayed improvements in Section~\ref{sec:2dbpp-allocation}, we compare the final programs discovered under the two replay quality measures. As in the main-text analysis, we use GPT-6-Luna and one completed run per measure and problem, with the same root programs and budgets for solver discovery and policy learning. Only the quality term changes from the top-five mean $Q_5$ in Eq.~\eqref{eq:q} to the best revealed score $Q_1$, while the request penalty remains unchanged. For each run, we select the solver with the highest development score and evaluate it directly on the test set.

For TSP and KP, this comparison uses 3,000 test instances per problem with $N\in\{50,100,200\}$. For OBPP, FSSP, and 2D-BPP, we use 30, 110, and 500 test instances, respectively. Table~\ref{tab:replay-quality-ablation} reports the best development score and test performance for each quality measure. $Q_5$ improves both metrics on TSP, FSSP, and 2D-BPP and matches $Q_1$ on KP at the displayed precision. These final results complement the higher delayed-improvement rates with $Q_5$ in Figure~\ref{fig:replay-quality-delayed}, supporting its use for guiding exploration toward better programs.

\begin{table}[!htbp]
\centering\small
\setlength{\tabcolsep}{6pt}
\begin{tabular}{@{}lrrlrr@{}}
\toprule
\multirow{2}{*}{Problem} & \multicolumn{2}{c}{Best dev. score$\uparrow$} & \multirow{2}{*}{Test metric} & \multicolumn{2}{c}{Test result} \\
 & $Q_1$ & $Q_5$ & & $Q_1$ & $Q_5$ \\
\midrule
TSP & 0.903628 & \textbf{0.911622} & Tour length$\downarrow$ & 9.7762 & \textbf{9.2079} \\
KP & \textbf{0.401890} & \textbf{0.401890} & Value$\uparrow$ & \textbf{39.25212} & \textbf{39.25212} \\
OBPP & \textbf{0.986612} & 0.984093 & Gap (\%)$\downarrow$ & \textbf{1.1238} & 1.1988 \\
FSSP & 0.549795 & \textbf{0.549904} & Gap (\%)$\downarrow$ & 0.1864 & \textbf{0.0412} \\
2D-BPP & 0.752726 & \textbf{0.755089} & Bins$\downarrow$ & 14.520 & \textbf{14.504} \\
\bottomrule
\end{tabular}
\caption{Comparison of replay quality terms $Q_1$ and $Q_5$ with GPT-6-Luna, using one run per quality term and problem. Test results use the solver with the highest development score in each run. Bold marks the better value within each problem and metric, including ties at displayed precision.}
\label{tab:replay-quality-ablation}
\end{table}

\FloatBarrier
\Needspace{6\baselineskip}
\subsection{2D-BPP Ablations and Costs}
\label{sec:2dbpp-cohort}
\paragraph{Shared paired-comparison protocol.}
All paired 2D-BPP comparisons use the same 500 BWMV instances with a 60-second budget per instance. For each method, $\bar{K}_i$ is the mean bin count across three runs on instance $i$. Each instance contributes one comparison, and a lower mean bin count is better. Ties are identified by exact equality of the integer bin-count sums across the three runs. We use two-sided exact sign tests of equal win and loss probabilities, excluding ties.

\paragraph{Paired \fixed{}/\dream{} comparison.}
\label{sec:2dbpp-fixed-dream-paired}
\begin{wraptable}{r}{0.48\textwidth}
\centering\small
\setlength{\tabcolsep}{4pt}
\resizebox{\linewidth}{!}{%
\begin{tabular}{@{}lrrr@{}}
\toprule
Model & W/T/L & Mean $\delta_i$ & $p$ \\
\midrule
GPT-6-Luna & 20/479/1 & 0.016 & $2.10\times10^{-5}$ \\
GPT-6-Sol & 13/486/1 & 0.011 & $0.002$ \\
\bottomrule
\end{tabular}
} 
\caption{Paired 2D-BPP comparison using three-run mean bin counts. W/T/L counts wins, ties, and losses for \dream{}. Mean $\delta_i$ reports bins saved per instance by \dream{}. The p-values are from two-sided exact sign tests excluding ties.
}
\label{tab:2dbpp-fixed-dream-paired}
\end{wraptable}

For GPT-6-Luna, \dream{} uses $\lambda=10^{-5}$. We define $\delta_i=\bar{K}_{i,\mathrm{\fixed{}}}-\bar{K}_{i,\mathrm{\dream{}}}$ as the bin-count reduction achieved by \dream{} on instance i. Positive, zero, and negative values give a \dream{} win, tie, and loss, respectively. Table~\ref{tab:2dbpp-fixed-dream-paired} shows more wins than losses for \dream{} with both backbones. Across the 500 instances, the total savings average 8 bins with GPT-6-Luna and 5.7 bins with GPT-6-Sol over three runs.

\dreamwrapclear
\paragraph{No replay protocol.}
No replay uses the same historical-evidence interface, history eligibility rule, and search budget ($BL=24$, $R=10$) as \dream{}. At each policy update, it makes a single LLM call to generate one policy and deploys it at $\alpha=0.5$ after syntax and interface checks. It performs no replay or reward-based ranking. \dream{} instead uses four proposal calls and replay-based selection of the policy–parameter pair, as described in Appendix~\ref{sec:contract}.

\paragraph{Paired replay comparison.}
\label{sec:replay-paired}
Across three runs, \dream{} at $\lambda=10^{-5}$ achieves a higher mean final development score than No replay (0.75451 versus 0.75327) with fewer search steps (189.7 versus 233.3). On the 500 test instances, the paired comparison records 15 wins, 483 ties, and 2 losses for \dream{}. A two-sided exact sign test on the 17 non-tied instances gives $p=0.00235$.

\par\medskip
\noindent\begin{minipage}[t]{0.48\textwidth}
\vspace{0pt}
\paragraph{Cost-penalty sensitivity.}
Table~\ref{tab:2dbpp-sensitivity} shows the effect of the cost penalty on development quality, search steps, and test bin counts. Among the evaluated settings, $10^{-5}$ achieves the highest development score and the lowest mean test bin count. Increasing $\lambda$ to $10^{-3}$ reduces mean search steps from 189.7 to 106.0 (44.1\%), with an increase of 0.010 bins per test instance.

\medskip
{
\centering\small
\setlength{\tabcolsep}{4pt}
\resizebox{\linewidth}{!}{%
\begin{tabular}{@{}lrrrr@{}}
\toprule
Method & $\lambda$ & Dev. score $\uparrow$ & Steps$\downarrow$ & Test bins $\downarrow$ \\
\midrule
\fixed{} & -- & 0.75290 & 240.0 & 14.51933 \\
\dream{} & $10^{-7}$ & 0.75417 & 204.7 & 14.51333 \\
\dream{} & $10^{-5}$ & \textbf{0.75451} & 189.7 & \textbf{14.50333} \\
\dream{} & $10^{-3}$ & 0.75209 & \textbf{106.0} & 14.51333 \\
\bottomrule
\end{tabular}
} \captionof{table}{Effect of the cost penalty $\lambda$ on 2D-BPP with GPT-6-Luna. Entries average three runs. Dev. score reports the final best development score. Steps counts discovery requests. Test bins reports the mean bin count over 500 test instances. Bold marks the best value in each metric.
}
\label{tab:2dbpp-sensitivity}
}

\medskip
\paragraph{Search trajectories.}
In Figure~\ref{fig:2dbpp-generation}, development requests are indexed in submission order, excluding root initialization. The horizontal axis begins at step 24 to highlight progress beyond the initial search phase. Figure~\ref{fig:2dbpp-attempts} shows that \dream{} at $\lambda=10^{-5}$ surpasses \fixed{}'s final mean development score by the third round, while using fewer search steps per round thereafter.

\end{minipage}\hfill
\begin{minipage}[t]{0.48\textwidth}
\vspace{0pt}
\centering
\includegraphics[width=\linewidth]{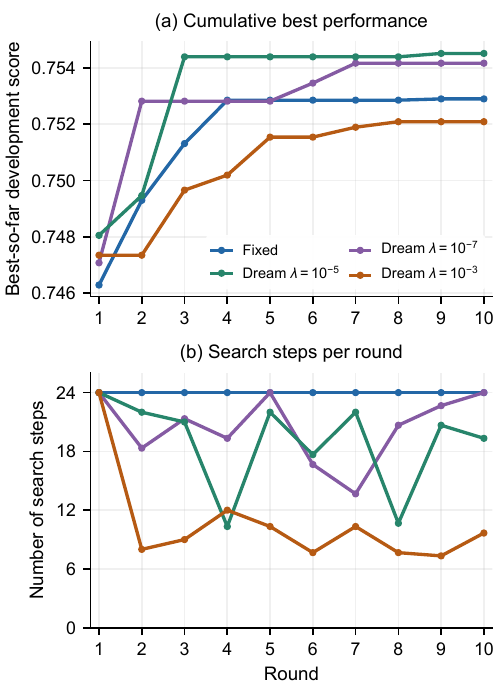}
\captionof{figure}{Round-level 2D-BPP development quality and search effort with GPT-6-Luna. (a) Cumulative best development score at the end of each round. (b) Search steps per round, including failures. Each point averages three runs.
}
\label{fig:2dbpp-attempts}
\end{minipage}
\par\medskip

\paragraph{Token costs.}
\begin{wrapfigure}{r}{0.48\textwidth}
\centering
\includegraphics[width=\linewidth]{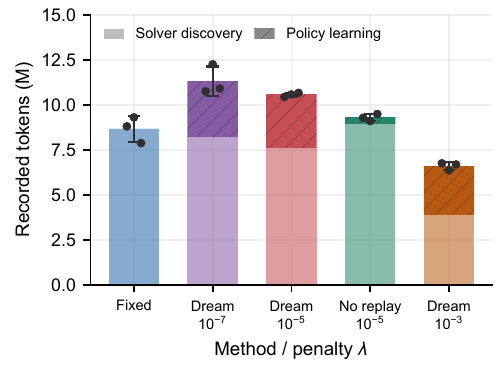}
\vspace{-2em}
\caption{Recorded token usage for 2D-BPP with GPT-6-Luna. Stacked bars show three-run means, split into algorithm discovery (unhatched) and policy learning (hatched). Whiskers indicate $\pm1$ sample standard deviation of total tokens. Dots show total tokens for individual runs.
}
\label{fig:token-usage}
\end{wrapfigure}

Figure~\ref{fig:token-usage} decomposes recorded token usage into solver discovery and policy learning. Across the three penalty settings, mean policy-learning usage ranges from 2.68 to 3.06 M tokens. Algorithm-discovery usage falls from 8.26 M at $\lambda=10^{-7}$ to 3.92 M at $\lambda=10^{-3}$. Most of the reduction therefore comes from solver discovery.

At $\lambda=10^{-3}$, \dream{} reduces discovery tokens by 54.7\% relative to \fixed{}. Adding 2.68 M policy-learning tokens brings its mean total to 6.61 M, compared with 8.67 M for \fixed{}. The reduction in discovery expenditure exceeds the policy-learning cost by 2.06 M tokens, yielding a net reduction of 23.8\%.

One cancelled call in a \fixed{} run has no recorded usage. As a sensitivity check, assigning it the 0.029 M tokens recorded for its successful retry increases the three-run \fixed{} mean by 0.11\%.

\dreamwrapclear
\subsection{Learned Policies in Real Search}
\label{sec:policy-trajectories}
We examine TSP and 2D-BPP search trajectories to illustrate how learned policies adapt branch continuation, revisiting, and stopping to observed progress. Both cases use GPT-6-Luna and show the ninth search round. Node labels indicate the order of search steps, and branch depth counts successive revisions from the root.

\begin{figure*}[!htbp]
\centering
\vspace{1em}
\includegraphics[width=\textwidth]{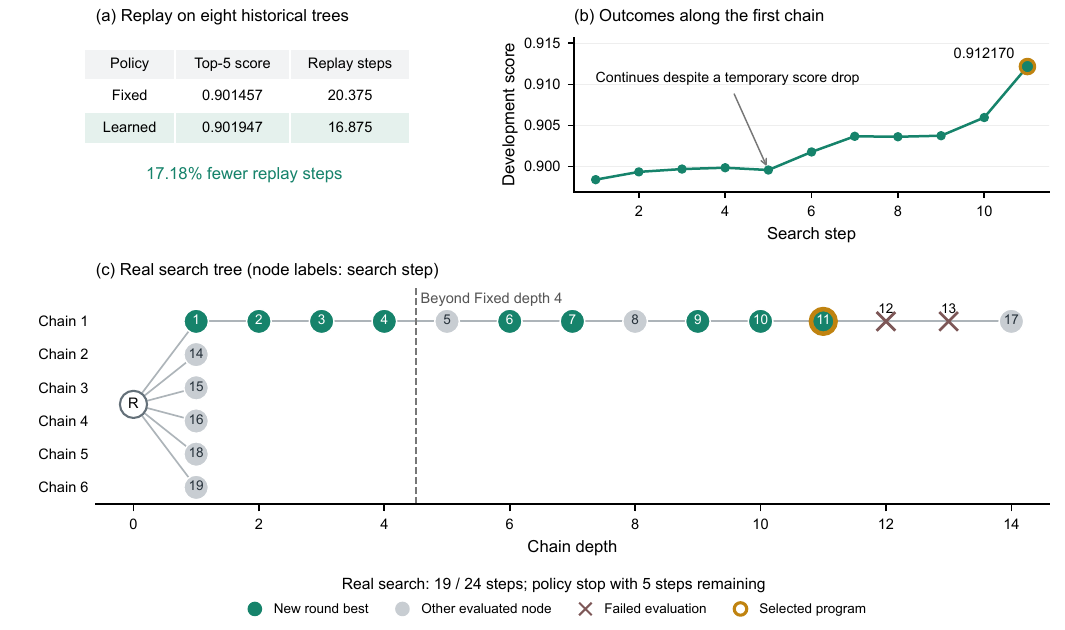}
\caption{TSP search round 9 with GPT-6-Luna. (a) Mean top-five score and replay steps on eight historical trees before deployment. (b) Development scores along the first branch up to the program selected for testing. (c) Complete search tree, with nodes labeled by search step. The policy stops after 19 of the 24 available steps.
}
\label{fig:tsp-policy-trajectory}
\end{figure*}

\paragraph{TSP: extending a productive branch and stopping early.}
Figure~\ref{fig:tsp-policy-trajectory} follows the policy in Listing~\ref{lst:policy-example}, deployed in round nine with $\alpha=0.9$. Before deployment, replay on eight historical trees yields a mean top-five score of 0.901947, compared with 0.901457 for \fixed{}. Mean replay expenditure falls from 20.375 to 16.875 steps per tree, a reduction of 17.18\%.

In real search, the policy extends the first branch beyond \fixed{}'s four-step depth limit. The development score falls from 0.899847 at depth four to 0.899567 at depth five, then recovers to 0.901768 at depth six. The policy's recent-gain and stagnation conditions permit continuation through this temporary decline. At depth eleven, the branch reaches 0.912170. This node contains the solver with the highest development score in the run, which is selected for testing.

After two failed evaluations on this branch, the policy explores three other branches before returning to the first at step 17. It then opens two further branches. With all six branches opened and no continuation exceeding its threshold, the policy stops after 19 search steps, leaving five steps unused.

\paragraph{2D-BPP: reallocating a full search budget.}
At $\lambda=10^{-5}$, the policy uses all 24 search steps but distributes them unevenly across branches (Figure~\ref{fig:2dbpp-policy-trajectory}). It ranks branch extensions using recent gains, stagnation, and quality gaps, and compares them with starting a new branch. The resulting branch lengths are $(8,6,1,3,5,1)$, whereas \fixed{} assigns four steps to each of six branches.

The first branch reaches 0.751209 at depth four (step 12) and improves to 0.753565 at depth five. After two lower-scoring revisions, the policy pauses this branch and allocates steps 16--19 elsewhere. Returning at step 20 produces a score of 0.755089 at depth eight, the highest development score in the run. The corresponding solver is selected for testing.

\FloatBarrier

\section{Programs Discovered by \dream{}}
\label{sec:discovered-programs}
We provide the \dream{} programs for each task. Each listing reproduces the complete evaluated source, with line wrapping for presentation only. TSP and KP include their construction wrappers. OBPP and FSSP expose learned components to the fixed controllers described in Appendix~\ref{sec:task-details}. The 2D-BPP program implements the full solver. Each listing shows a single selected program; its performance below refers to that program. Table~\ref{tab:constructive} averages three runs for GPT-6-Luna TSP, and Table~\ref{tab:fullsolver} averages three runs for each 2D-BPP method and backbone.

\subsection{Constructive TSP}
\label{sec:code-tsp}
The GPT-6-Luna program in Listing~\ref{lst:dream-tsp} combines short lookahead with residual minimum-spanning-tree and greedy completion costs. It solves the final ten or fewer unvisited cities by subset dynamic programming. Its mean gap to LKH is 11.00\% over all 4,015 test instances, comprising 4,000 synthetic instances and 15 TSPLIB instances.

\begin{lstlisting}[ style=dreamprogram, caption={\dream{}'s constructive TSP program.}, label={lst:dream-tsp} ]
import numpy as np
import math

def select_next_node(current_node, destination_node, unvisited_nodes, distance_matrix):
    nodes = sorted(unvisited_nodes)
    if len(nodes) == 1:
        return int(nodes[0])

    # Solve the small endgame exactly, including the final return to the start.
    if len(nodes) <= 10:
        count = len(nodes)
        full_mask = (1 << count) - 1
        memo = {}

        def tail_cost(last_index, remaining_mask):
            key = (last_index, remaining_mask)
            if key in memo:
                return memo[key]
            if remaining_mask == 0:
                return float(distance_matrix[nodes[last_index], destination_node])
            best = float("inf")
            for next_index in range(count):
                bit = 1 << next_index
                if remaining_mask & bit:
                    cost = float(distance_matrix[nodes[last_index], nodes[next_index]])
                    cost += tail_cost(next_index, remaining_mask ^ bit)
                    if cost < best:
                        best = cost
            memo[key] = best
            return best

        best_cost = float("inf")
        best_city = nodes[0]
        for index, city in enumerate(nodes):
            cost = float(distance_matrix[current_node, city])
            cost += tail_cost(index, full_mask ^ (1 << index))
            if cost < best_cost:
                best_cost = cost
                best_city = city
        return int(best_city)

    current_distances = np.asarray(distance_matrix[current_node, nodes], dtype=float)
    order = np.argsort(current_distances, kind="stable")
    shortlist_size = min(len(nodes), max(3, int(math.ceil(math.sqrt(len(nodes))))))
    candidates = [nodes[int(i)] for i in order[:shortlist_size]]

    def residual_mst_cost(vertices):
        count = len(vertices)
        if count <= 1:
            return 0.0
        used = [False] * count
        best_edge = [float("inf")] * count
        best_edge[0] = 0.0
        total = 0.0
        for _ in range(count):
            chosen = -1
            chosen_cost = float("inf")
            for i in range(count):
                if not used[i] and best_edge[i] < chosen_cost:
                    chosen = i
                    chosen_cost = best_edge[i]
            if chosen < 0:
                break
            used[chosen] = True
            total += chosen_cost
            source = vertices[chosen]
            for i in range(count):
                if not used[i]:
                    edge = float(distance_matrix[source, vertices[i]])
                    if edge < best_edge[i]:
                        best_edge[i] = edge
        return total

    def greedy_tail_cost(start, vertices):
        remaining = set(vertices)
        last = start
        total = 0.0
        while remaining:
            nxt = min(remaining, key=lambda node: (float(distance_matrix[last, node]), node))
            total += float(distance_matrix[last, nxt])
            remaining.remove(nxt)
            last = nxt
        total += float(distance_matrix[last, destination_node])
        return total

    best_key = None
    best_city = candidates[0]
    for city in candidates:
        first = float(distance_matrix[current_node, city])
        remaining = [node for node in nodes if node != city]

        if remaining:
            second = min(float(distance_matrix[city, node]) for node in remaining)
            next_city = min(remaining, key=lambda node: (float(distance_matrix[city, node]), node))
            after_next = [node for node in remaining if node != next_city]
            if after_next:
                third = min(float(distance_matrix[next_city, node]) for node in after_next)
            else:
                third = float(distance_matrix[next_city, destination_node])
        else:
            second = third = 0.0

        residual_vertices = list(remaining)
        if destination_node not in residual_vertices:
            residual_vertices.append(destination_node)
        global_tail = residual_mst_cost(residual_vertices)
        completion_tail = greedy_tail_cost(city, remaining)
        score = first + 0.12 * second + 0.04 * third + 0.12 * global_tail + 0.18 * completion_tail
        key = (score, first, city)
        if best_key is None or key < best_key:
            best_key = key
            best_city = city
    return int(best_city)

def solve(instance, time_budget, seed):
    points = np.asarray(instance["points"], dtype=float)
    n = len(points)
    distance_matrix = np.sqrt(np.sum((points[:, None, :] - points[None, :, :]) ** 2, axis=-1))
    tour = [0]
    unvisited = set(range(1, n))
    while unvisited:
        node = select_next_node(tour[-1], 0, unvisited.copy(), distance_matrix.copy())
        if isinstance(node, (bool, np.bool_)) or not isinstance(node, (int, np.integer)) or node not in unvisited:
            raise ValueError("select_next_node must return an unvisited integer city")
        node = int(node)
        tour.append(node)
        unvisited.remove(node)
    return tour
\end{lstlisting}

\subsection{Constructive Knapsack}
\label{sec:code-kp}
The GPT-6-Luna selector in Listing~\ref{lst:dream-kp} uses a beam search ordered by value density and guided by an optimistic completion bound. It returns the first item of the best retained partial solution, and the constructor repeats this decision as capacity decreases. Its mean value is 52.1798 over all 4,000 test instances at the four reported item counts.

\begin{lstlisting}[ style=dreamprogram, caption={\dream{}'s constructive knapsack program.}, label={lst:dream-kp} ]
import numpy as np

def select_next_item(remaining_capacity, weights, values):
    n = len(weights)
    if n == 0:
        raise ValueError("No items are available")

    capacity = float(remaining_capacity)
    feasible = np.flatnonzero(weights <= capacity)
    if not len(feasible):
        # Tolerance-feasible fallback required when no strictly feasible item exists.
        return int(np.argmin(weights))

    # Process promising items first so the beam explores useful combinations early.
    density = values / np.maximum(weights, 1e-15)
    order = np.argsort(-density, kind="mergesort")
    sorted_weights = weights[order]
    sorted_values = values[order]
    sorted_density = density[order]

    # Each state is (used weight, collected value, first selected current index).
    # A beam bound adds the fractional-knapsack value available from later items.
    beam = [(0.0, 0.0, -1)]
    beam_width = 128
    for position in range(n):
        item_weight = float(sorted_weights[position])
        item_value = float(sorted_values[position])
        current_index = int(order[position])
        candidates = []
        for used_weight, total_value, first_item in beam:
            candidates.append((used_weight, total_value, first_item))
            new_weight = used_weight + item_weight
            if new_weight <= capacity:
                new_first = current_index if first_item < 0 else first_item
                candidates.append((new_weight, total_value + item_value, new_first))

        next_density = float(sorted_density[position + 1]) if position + 1 < n else 0.0
        remaining_total_weight = float(np.sum(sorted_weights[position + 1:]))
        ranked = []
        for used_weight, total_value, first_item in candidates:
            residual = max(0.0, capacity - used_weight)
            # The fractional bound is inexpensive and admissible for the remaining items.
            optimistic_extra = min(residual, remaining_total_weight) * next_density
            ranked.append((-(total_value + optimistic_extra), -total_value, used_weight, first_item, used_weight, total_value, first_item))
        ranked.sort()
        beam = [(row[4], row[5], row[6]) for row in ranked[:beam_width]]

    best = min(beam, key=lambda state: (-state[1], state[0], state[2] if state[2] >= 0 else n))
    if best[2] < 0:
        return int(feasible[0])
    return int(best[2])

def solve(instance, time_budget, seed):
    weights = np.asarray(instance["weights"], dtype=float)
    values = np.asarray(instance["values"], dtype=float)
    remaining_capacity = float(instance["capacity"])
    epsilon = float(instance["feasibility_epsilon"])
    indices = np.arange(len(weights))
    selected = []
    while len(weights) and remaining_capacity + epsilon > float(weights.min()):
        item = select_next_item(remaining_capacity=remaining_capacity, weights=weights.copy(), values=values.copy())
        if isinstance(item, (bool, np.bool_)) or not isinstance(item, (int, np.integer)) or not 0 <= item < len(weights):
            raise ValueError("select_next_item must return an integer index in the current arrays")
        item = int(item)
        if not remaining_capacity + epsilon > weights[item]:
            raise ValueError("Selected item exceeds remaining capacity")
        selected.append(int(indices[item]))
        remaining_capacity -= weights[item]
        weights = np.delete(weights, item)
        values = np.delete(values, item)
        indices = np.delete(indices, item)
    return selected
\end{lstlisting}

\subsection{Online Bin Packing}
\label{sec:code-obpp}
The GPT-6-Luna scoring rule in Listing~\ref{lst:dream-obpp} penalizes small positive residual capacity and rewards exact fits. It uses 1,297.0 bins on average over the 30 test instances, with a mean gap of 1.199\% to the lower bound. The benchmark supplies \texttt{\_obpp\_pack}, the fixed controller that applies these scores to feasible bins.

\begin{lstlisting}[ style=dreamprogram, caption={\dream{}'s online bin-packing scoring rule and controller entry point.}, label={lst:dream-obpp} ]
import numpy as np

def score(item, bins):
    remaining = np.asarray(bins, dtype=float) - float(item)
    scores = -remaining.copy()

    # Future items have a typical size near the Weibull scale (45).
    # Smoothly discourage positive scraps smaller than that size.
    scrap_limit = 45.0
    tiny_scrap = (remaining > 1e-9) & (remaining < scrap_limit)
    deficit = (scrap_limit - remaining[tiny_scrap]) / scrap_limit
    scores[tiny_scrap] -= 3.0 * scrap_limit * deficit * deficit

    # Exact fits avoid consuming capacity that could serve later items.
    scores[remaining <= 1e-9] += 100.0
    return scores

def solve(instance, time_budget, seed):
    return _obpp_pack(instance, seed, score)
\end{lstlisting}

\subsection{Flow-Shop Scheduling}
\label{sec:code-fssp}
The GPT-6-Luna program in Listing~\ref{lst:dream-fssp} traces a critical path to prioritize jobs and machines for perturbation. Its mean gap to the upper bounds is 0.0412\% over 110 test instances. The benchmark supplies \texttt{\_fssp\_gls}, the fixed NEH-initialized guided local search controller that calls \texttt{get\_matrix\_and\_jobs}.

\begin{lstlisting}[ style=dreamprogram, caption={\dream{}'s flow-shop perturbation rule and controller entry point.}, label={lst:dream-fssp} ]
import numpy as np


def get_matrix_and_jobs(current_sequence, time_matrix, m, n):
    times = np.asarray(time_matrix, dtype=float)
    jobs_count, machines = times.shape
    if jobs_count == 0 or machines == 0:
        return times.copy(), []

    # Keep one valid occurrence of each job in sequence order.
    seq = []
    seen = set()
    for value in current_sequence:
        job = int(value)
        if 0 <= job < jobs_count and job not in seen:
            seq.append(job)
            seen.add(job)
    if not seq:
        seq = list(range(jobs_count))

    # Completion times support a backward trace of one critical path.
    completion = np.zeros((len(seq), machines), dtype=float)
    for i, job in enumerate(seq):
        for k in range(machines):
            above = completion[i - 1, k] if i > 0 else 0.0
            left = completion[i, k - 1] if k > 0 else 0.0
            completion[i, k] = max(above, left) + times[job, k]

    path_counts = np.zeros(jobs_count, dtype=float)
    critical_machines = np.zeros(machines, dtype=float)
    path_positions = np.zeros(len(seq), dtype=bool)
    i, k = len(seq) - 1, machines - 1
    while i >= 0 and k >= 0:
        job = seq[i]
        path_positions[i] = True
        path_counts[job] += 1.0
        critical_machines[k] += 1.0
        if i == 0 and k == 0:
            break
        above = completion[i - 1, k] if i > 0 else -1.0
        left = completion[i, k - 1] if k > 0 else -1.0
        if above >= left:
            i -= 1
        else:
            k -= 1

    # Add a modest boundary signal for jobs directly adjacent to critical
    # operations; these jobs often determine whether a critical block can move.
    boundary = np.zeros(jobs_count, dtype=float)
    for pos, on_path in enumerate(path_positions):
        if on_path:
            if pos > 0 and not path_positions[pos - 1]:
                boundary[seq[pos - 1]] += 0.5
            if pos + 1 < len(seq) and not path_positions[pos + 1]:
                boundary[seq[pos + 1]] += 0.5

    total_load = np.sum(times, axis=1)
    critical_load = times @ critical_machines
    mean_load = float(np.mean(critical_load)) + 1e-12
    priority = (2.5 * path_counts + boundary + critical_load / mean_load
                + 0.15 * total_load / (float(np.mean(total_load)) + 1e-12))

    count = min(5, jobs_count)
    pool_size = min(jobs_count, max(count, min(16, 3 * count)))
    order = sorted(range(jobs_count), key=lambda j: (priority[j], path_counts[j], total_load[j]), reverse=True)
    pool = np.asarray(order[:pool_size], dtype=int)
    weights = np.asarray([1.0 + max(0.0, priority[j]) for j in pool], dtype=float)
    weights = np.maximum(weights, 1e-9)
    probabilities = weights / np.sum(weights)
    chosen = np.random.choice(pool, size=count, replace=False, p=probabilities).tolist()

    # Concentrate perturbation on machines appearing on the path, with a small
    # shoulder on neighboring machines to expose adjacent bottleneck effects.
    focus = critical_machines.copy()
    for machine in range(machines):
        amount = critical_machines[machine]
        if amount > 0.0:
            if machine > 0:
                focus[machine - 1] += 0.25 * amount
            if machine + 1 < machines:
                focus[machine + 1] += 0.25 * amount
    if not np.any(focus > 0.0):
        focus[-1] = 1.0
    focus /= max(float(np.max(focus)), 1e-12)

    modified = times.copy()
    for job in chosen:
        # Jobs with greater path participation receive a stronger perturbation.
        job_strength = 0.85 + 0.15 * min(1.0, path_counts[int(job)] / max(1.0, float(np.max(path_counts))))
        factor = float(np.random.uniform(1.15, 1.45)) * job_strength
        scale = np.maximum(focus, 0.10 * (focus > 0.0))
        modified[int(job), :] *= 1.0 + (factor - 1.0) * scale

    modified = np.nan_to_num(modified, nan=0.0, posinf=1.0e12, neginf=0.0)
    modified = np.maximum(modified, 0.0)
    return modified, [int(job) for job in chosen]


def solve(instance, time_budget, seed):
    return _fssp_gls(instance, time_budget, seed, get_matrix_and_jobs)
\end{lstlisting}

\subsection{Complete Two-Dimensional Bin Packing}
\label{sec:code-2dbpp}
The full solver in Listing~\ref{lst:dream-2dbpp} was discovered with GPT-6-Sol at $\lambda=0$. It combines free-rectangle placement with randomized item orders and partial repacking. Accepted layouts are checked for feasibility, and a valid fallback is maintained throughout the search. It uses 7,240 bins over the 500 BWMV instances, or 14.480 bins per instance.

\begin{lstlisting}[ style=dreamprogram, caption={\dream{}'s complete two-dimensional bin-packing solver.}, label={lst:dream-2dbpp} ]
import random
import time


def _valid(placements, items, width, height):
    if not isinstance(placements, list) or len(placements) != len(items):
        return False
    bins = {}
    for i, placement in enumerate(placements):
        if not isinstance(placement, (list, tuple)) or len(placement) != 3:
            return False
        if any(type(value) is not int or value < 0 for value in placement):
            return False
        bid, x, y = placement
        w, h = items[i]
        if x + w > width or y + h > height:
            return False
        for ox, oy, ow, oh in bins.get(bid, ()):
            if x < ox + ow and ox < x + w and y < oy + oh and oy < y + h:
                return False
        bins.setdefault(bid, []).append((x, y, w, h))
    return True


def _split_free(free, x, y, w, h):
    pieces = []
    right = x + w
    top = y + h
    for fx, fy, fw, fh in free:
        fr = fx + fw
        ft = fy + fh
        if right <= fx or fr <= x or top <= fy or ft <= y:
            pieces.append((fx, fy, fw, fh))
            continue
        if x > fx:
            pieces.append((fx, fy, x - fx, fh))
        if right < fr:
            pieces.append((right, fy, fr - right, fh))
        if y > fy:
            pieces.append((fx, fy, fw, y - fy))
        if top < ft:
            pieces.append((fx, top, fw, ft - top))
    result = []
    for i, rect in enumerate(pieces):
        ax, ay, aw, ah = rect
        contained = False
        for j, other in enumerate(pieces):
            if i == j:
                continue
            bx, by, bw, bh = other
            if bx <= ax and by <= ay and ax + aw <= bx + bw and ay + ah <= by + bh:
                if rect != other or j < i:
                    contained = True
                    break
        if not contained:
            result.append(rect)
    return result


def _contact(x, y, w, h, rectangles, width, height):
    value = (h if x == 0 else 0) + (h if x + w == width else 0)
    value += (w if y == 0 else 0) + (w if y + h == height else 0)
    for ox, oy, ow, oh in rectangles:
        if x + w == ox or ox + ow == x:
            value += max(0, min(y + h, oy + oh) - max(y, oy))
        if y + h == oy or oy + oh == y:
            value += max(0, min(x + w, ox + ow) - max(x, ox))
    return value


def _build(order, kept, base, items, width, height, style, deadline):
    placements = [None] * len(items)
    bins = []
    remap = {}
    for i in kept:
        if time.monotonic() >= deadline:
            return None
        old_bid, x, y = base[i]
        if old_bid not in remap:
            remap[old_bid] = len(bins)
            bins.append([[(0, 0, width, height)], 0, []])
        bid = remap[old_bid]
        w, h = items[i]
        bins[bid][0] = _split_free(bins[bid][0], x, y, w, h)
        bins[bid][1] += w * h
        bins[bid][2].append((x, y, w, h))
        placements[i] = [bid, x, y]
    for i in order:
        if time.monotonic() >= deadline:
            return None
        w, h = items[i]
        area = w * h
        best_key = None
        options = []
        for bid, (free, used, rectangles) in enumerate(bins):
            if time.monotonic() >= deadline:
                return None
            for fx, fy, fw, fh in free:
                if w > fw or h > fh:
                    continue
                waste = fw * fh - area
                short = min(fw - w, fh - h)
                if style == 0:
                    key = (-used, waste, short)
                elif style == 1:
                    key = (waste, -used, short)
                elif style == 2:
                    key = (short, -used, waste)
                else:
                    key = (-used, short, waste)
                if best_key is None or key < best_key:
                    best_key = key
                    options = []
                if key == best_key:
                    for x in (fx, fx + fw - w):
                        for y in (fy, fy + fh - h):
                            options.append((bid, x, y))
        if not options:
            bid = len(bins)
            bins.append([[(0, 0, width, height)], 0, []])
            x = y = 0
        else:
            bid, x, y = min(
                options,
                key=lambda p: (-_contact(p[1], p[2], w, h, bins[p[0]][2], width, height), p[2], p[1], p[0]),
            )
        bins[bid][0] = _split_free(bins[bid][0], x, y, w, h)
        bins[bid][1] += area
        bins[bid][2].append((x, y, w, h))
        placements[i] = [bid, x, y]
    return placements, [entry[1] for entry in bins]


def _order(indices, items, mode, rng, noise):
    priorities = []
    for i in indices:
        w, h = items[i]
        if mode == 0:
            priority = w * h
        elif mode == 1:
            priority = max(w, h)
        elif mode == 2:
            priority = w + h
        elif mode == 3:
            priority = w
        elif mode == 4:
            priority = h
        else:
            priority = min(w, h)
        if noise:
            priority *= 0.65 + 0.70 * rng.random()
        priorities.append((-priority, i))
    priorities.sort()
    return [i for _, i in priorities]


def solve(instance, time_budget, seed, lower_bound=0):
    start = time.monotonic()
    width, height = instance['bin_size']
    items = instance['items']
    count = len(items)
    if count == 0:
        return []
    fallback = [[i, 0, 0] for i in range(count)]
    if not _valid(fallback, items, width, height):
        raise ValueError('No feasible single-item-bin fallback')
    best = fallback
    areas = [w * h for w, h in items]
    best_numerator = sum(area * area for area in areas)
    best_bins = count
    rng = random.Random(seed)
    deadline = start + max(0.0, float(time_budget)) * 0.85
    iteration = 0
    while time.monotonic() < deadline:
        mode = iteration % 6
        style = iteration % 4 if iteration < 16 else rng.randrange(4)
        if iteration >= 16 and iteration % 5 != 0:
            occupied = {}
            for i, placement in enumerate(best):
                bid = placement[0]
                occupied[bid] = occupied.get(bid, 0) + areas[i]
            ranked = sorted(occupied, key=lambda bid: (occupied[bid], bid))
            pivot = rng.choice(ranked[:min(4, len(ranked))])
            selected = {pivot}
            target = min(len(ranked), 1 + rng.randrange(4))
            others = [bid for bid in ranked if bid != pivot]
            rng.shuffle(others)
            selected.update(others[:target - 1])
            kept = [i for i in range(count) if best[i][0] not in selected]
            removed = [i for i in range(count) if best[i][0] in selected]
            order = _order(removed, items, mode, rng, True)
            result = _build(order, kept, best, items, width, height, style, deadline)
        else:
            order = _order(range(count), items, mode, rng, iteration >= 6)
            result = _build(order, (), best, items, width, height, style, deadline)
        if result is None:
            break
        candidate, used_areas = result
        numerator = sum(area * area for area in used_areas)
        bin_count = len(used_areas)
        if numerator * best_bins > best_numerator * bin_count:
            if _valid(candidate, items, width, height):
                best = candidate
                best_numerator = numerator
                best_bins = bin_count
        iteration += 1
    return best if _valid(best, items, width, height) else fallback
\end{lstlisting}
\end{document}